\documentclass{article}

\usepackage{natbib}

\usepackage[a4paper,margin=3.0cm,foot=2.9cm]{geometry}
\usepackage{hyperref}
\usepackage[detect-weight=true]{siunitx}
\usepackage{booktabs}
\usepackage{listings}
\usepackage{tikz}
\usetikzlibrary{positioning, shapes.geometric, shapes.arrows}
\usepackage[percent]{overpic} 
\usepackage{makecell} 
\usepackage{amsmath}
\usepackage{setspace}

\newcommand{\KEYWORD}[1]{%
  \vspace{1em}%
  \noindent\textbf{Keywords:} #1%
  \vspace{1em}%
}

\usepackage{float}
\floatstyle{plain}
\newfloat{listing}{htbp}{lop}
\floatname{listing}{Listing}

\newenvironment{ttquotesmall}
 {\quote\ttfamily\raggedright\small}
 {\endquote}
 
\begin{document}%

\title{Curating art exhibitions using machine learning}

\author{Eurico Covas\\  
	e-mail: \href{mailto:eurico.covas@mail.com}{eurico.covas@mail.com}\\
	Collaborator, Instituto de Astrof\'isica e Ci\^encias do Espa\c{c}o (IA), Portugal\\
	and\\ 
    e-mail: \href{mailto:eurico.covas@edu.ulisboa.pt}{eurico.covas@edu.ulisboa.pt}\\
	Student, Faculdade de Belas-Artes da Universidade de Lisboa, Portugal
	}
	
\maketitle
\thispagestyle{plain}    
\pagestyle{plain}        
	

\begin{abstract}
Here we present a series of artificial models -- a total of four related models -- based on machine learning techniques that attempt to learn from existing exhibitions which have been curated by human experts, in order to be able to do  similar curatorship work.  
Out of our four artificial intelligence models, three achieve a reasonable ability at imitating these various curators responsible for all those exhibitions, with various degrees of precision and curatorial coherence. In particular, we can conclude two key insights: first, that there is sufficient information in these exhibitions to construct an artificial intelligence model that replicates past exhibitions with an accuracy well above random choices; and second, that using feature engineering and carefully designing the architecture of modest size models can make them almost as good as those using the so-called large language models such as GPT in a brute force approach. 

\end{abstract}

\KEYWORD{art curation; machine learning; embeddings; large language models (LLMs)}


\onehalfspacing

\section{Introduction}\label{introduction}

Artificial intelligence (AI) and its applications are currently a hot topic in academia as well as in the news, particularly with the announcement of the 2024 Nobel Prize in Physics, which was attributed to work by researchers  in AI \citep{nobel2024press}. In the area of art, AI has already been applied in several research fields, including: generating art (images) in any style one wants \citep{gatys2016image}; generating images from pure text \citep{ramesh2021zero, esser2021taming, radford2021learning};
translating paintings to possible corresponding real-world images (or vice versa) and other artistic styles \citep{zhu2017unpaired};
discovering similar patterns in disparate paintings \citep{saleh2016toward, shen2019discovering}; clustering similar artworks \citep{hamilton2021mosaic, castellano2022deep}; style/artistic school and artist identification/classification \citep{keren2002painter, shamir2010impressionism, shamir2012computer, cetinic2013automated, saleh2015large, moma_identifying_art_2025};
painting to text/semantic knowledge extraction \citep{vinyals2015show}; aesthetic evaluation and artistic opinion extraction \citep{cetinic2019deep};
and museum/exhibition curation and artwork selection/recommendation \citep{he2016vista, krysa2019next, messina2019content, messina2020curatornet, bowen2020computational, fosset2022docent, yilma2023elements, ohm2023algorithmic, nasher2023act, schaerf2023ai, 10.1162/leon_a_02561, von2024machine, srinivasan2024see, newsobserver2025, li2025enhanced}.

Regarding the last field of research, museum and exhibition curation using AI, most of the work seems to focus on either finding similar artworks to a given (seed) artwork, or using a prompt (e.g., for the OpenAI GPT-4 \citep{DBLP:journals/corr/abs-2303-08774}), to select appropriate artworks. Here we propose a different and modest approach that does not seem to have been  explored yet. We propose to use existing collections of actual past art exhibitions \citep[see e.g.,][and others]{artlas2025, univie2025, moma2025, nga2025, metmuseum_openaccess}, to create a recommender system \citep[for an example of recommender systems see][]{hu2008collaborative} using AI that imitates what human curators do. That is, we propose to teach AI to learn from past curatorship. The closest of the above-mentioned works to our proposal is the one by the Nasher Museum of Art at Duke University \citep{nasher2023act}, which used OpenAI's ChatGPT platform to curate exhibitions. The main difference with respect to our work here is that our modelling is absolutely restricted to a given set of artworks (therefore no hallucinations giving artworks outside that set), and more importantly and crucially, we have made the models learn from past exhibitions using those artworks and only learn from that set of information. 

This is how we propose to train our models.
The training set would consist of the input pairs $(x,y)$, where $x$ is the embedding vector \citep[see e.g.,][on text-to-vector embeddings]{kenton2019bert} of the exhibition title and main text, while $y$ is made of the features extracted from the images of the artworks exhibited. Optionally, one could add to $y$ the embedding vector of the caption of artworks that were selected by the human curators, if available.\footnote{We could consider also using the information on who was the human curator, for a more nuanced system, that could recognise and create an exhibition on the style of a certain human curator, but given the scarcity of information available, that is wildly optimistic.} After the training phase, users could input on the test phase, a title and a description of a hypothetical exhibition, and the  system would choose the appropriate artworks to include, and would choose only from the given  set of artworks.

\section{Dataset}

Although there are several datasets available  \citep[see e.g.,][and others]{artlas2025, univie2025, moma2025, nga2025, metmuseum_openaccess}, we have decided to restrict ourselves to the Metropolitan Museum of Art in New York (hereafter Met Museum), as it had two essential ingredients for our intended research project, namely: an extensive and well formatted dataset on past exhibitions and their works of art, and the corresponding freely available dataset on details and tags associated with those works of art, i.e., their metadata. Furthermore, the merging of databases could have proven a very complex task, as these sets all have different metadata associated to the artworks, and the data formats come in a large variety.

On the first part of this dataset, we have focused on the data from the Met Museum's website \citep{metmuseum_past_exhibitions}, and extracted directly the full set of archived exhibitions and where available, its corresponding works of art (or rather the object id, an integer that clearly identifies the artwork within the Met Museum's associated artwork database). This dataset included a total of \num{1636} exhibitions from 2000 to 2025 inclusive, although these included duplicated exhibitions, as some of those encompass several years. Also, some exhibitions did not contain the list of object ids to map to works of art and/or only contained objects temporarily loaned to the Met Museum and therefore not in the database of objects. After removing those and other exhibitions we could not use, we end up with a much lower value of \num{236} exhibitions. We depict in Table \ref{exhibitions_table} a summary of that data, which corresponds to a download on the date of 24 February 2025 (naturally, the Met Museum may have released more webpages with newer exhibitions, but we have frozen our data on this date). 

\begin{table}[htb]
\centering
\scriptsize
\begin{tabular}{c|c|c|c|c}
\hline
\textbf{Year} & \textbf{Total Exhibitions} & \textbf{Exhibitions with Object IDs} & \textbf{Works of Art Found} & \textbf{Word Count} \\
\hline
2000 & 55 & 6  & \num{168  } & \num{5911 }	\\
2001 & 59 & 11 & \num{403  } & \num{6554 }	\\
2002 & 54 & 11 & \num{254  } & \num{6242 }	\\
2003 & 56 & 9  & \num{371  } & \num{5911 }	\\
2004 & 57 & 6  & \num{342  } & \num{6052 }	\\
2005 & 49 & 4  & \num{239  } & \num{5544 }	\\
2006 & 49 & 5  & \num{308  } & \num{6147 }	\\
2007 & 51 & 16 & \num{847  } & \num{6019 }	\\
2008 & 57 & 19 & \num{1203 } & \num{6947 }	\\
2009 & 61 & 18 & \num{863  } & \num{6905 }	\\
2010 & 61 & 16 & \num{1307 } & \num{6246 }	\\
2011 & 68 & 24 & \num{1574 } & \num{7246 }	\\
2012 & 65 & 30 & \num{2661 } & \num{7373 }	\\
2013 & 79 & 39 & \num{2863 } & \num{9301 }	\\
2014 & 94 & 53 & \num{2827 } & \num{9832 }	\\
2015 & 90 & 53 & \num{2741 } & \num{8663 }	\\
2016 & 95 & 58 & \num{3646 } & \num{9103 }	\\
2017 & 95 & 2  & \num{111  } & \num{9694 }	\\
2018 & 78 & 2  & \num{129  } & \num{8222 }	\\
2019 & 74 & 2  & \num{129  } & \num{7462 }	\\
2020 & 56 & 5  & \num{276  } & \num{5876 }	\\
2021 & 58 & 26 & \num{2042 } & \num{9727 }	\\
2022 & 56 & 47 & \num{3711 } & \num{12796}	\\
2023 & 60 & 47 & \num{3483 } & \num{14078}	\\
2024 & 50 & 36 & \num{2986 } & \num{10434}	\\
2025 & 9  & 8  & \num{736  } & \num{1590 }	\\
\hline
\textbf{Total} & \textbf{1636} & \textbf{553} & \textbf{\num{36220}} & \textbf{\num{199875}} \\
\hline
\shortstack{\textbf{Total after} \\ \textbf{removal of} \\ \textbf{multi-year } \\ \textbf{exhibitions}} & \textbf{1009} & \textbf{338} & \textbf{\num{20172}} &\textbf{\num{123349}}\\
\hline
\shortstack{\textbf{Total after} \\ \textbf{removal of} \\ \textbf{artworks not} \\ \textbf{on Met Museum}} & \textbf{\,} & \textbf{236} & \textbf{\num{10470}} &\textbf{\num{26388}} \\
\hline
\end{tabular}
\caption{Exhibition statistics per year for the downloaded exhibitions from the Metropolitan Museum of Art of New York,  which corresponds to a download on the date of 24 February 2025. It shows the total number of exhibitions per year, the number of those that have the artworks explicitly stated (using Object IDs that match the artwork database from the Met Museum) and the number of artworks found. As some exhibitions run across year ends, we also show the corresponding total all years' value of the number of exhibitions that have known Object IDs and the corresponding lower number of actual artworks in that unique set of exhibitions. Furthermore, we also show those numbers when one removes exhibitions that encompass only artworks not on the Met Museum's database, i.e., exhibitions that are made of loaned artworks. The actual usable number of exhibitions for our research project is therefore only 236, and the number of artworks a total of \num{10470} (with possible repeats across exhibitions). The number of actual unique artworks is  \num{9289}. Finally, we show the number of words in each year for all the titles and descriptions for all exhibitions within that year, and the total number of words for all exhibitions with and without duplication, and for the actual exhibitions we shall use for our modelling. For comparison, it is speculated that the OpenAI's \texttt{gpt-4o-mini} model uses several billion words for its training.}
\label{exhibitions_table}
\end{table}


The exhibition dataset, consisting of a grand total of 236 exhibitions after removal of multi-year duplicates and made of exhibitions for which we have object ids in the corresponding Met Museum's artworks object id to metadata file, has the following format, as depicted in Listing \ref{all_exhibitions_json}.
For each exhibition, we have the title of the exhibition and the description of the exhibition (called \texttt{overview\_text}), all in English and with 
the replacement of any diacritics (accents, umlauts, tildes, etc.) and non-ASCII characters, and then we have a variable number of \texttt{object\_ids}, which identify the artworks within the Met Museum.\footnote{
Note that this id is not a worldwide id, but an internal Met Museum id. There are some attempts to catalogue and identify all artworks with a unique worldwide id, e.g.,: Wikidata  \texttt{Qxxxxx} ids --- ``Les Demoiselles d'Avignon'' by Pablo Picasso has a unique id \texttt{Q910199}; the
 International Image Interoperability Framework \citep{iiif2024}; the Europeana project \citep{europeana2024}; among many others.} For each
\texttt{object\_id}, we have a series of metadata elements, which can be a single string, a list of strings or an empty list with no elements in it.

\begin{listing}[htb]
\begin{lstlisting}
{
    "exhibitions": [
        {
            "title": "Sculpture and Decorative Arts of the Spanish Renaissance",
            "overview_text": "The Metropolitan Museum of Art's small but excellent collection of Spanish polychrome sculpture, including sacred reliefs and freestanding carved figures once housed in the churches of Spain, is displayed in the gallery adjacent to the newly reopened Velez Blanco Patio. The selection, which displays the unique blending of early western European and Islamic stylistic and technical influences, emphasizes the diversity in the material culture of Renaissance Spain after the Catholic reconquest by Ferdinand and Isabella.",
            "object_ids": {
                "187702": {
                    "Department": "European Sculpture and Decorative Arts",
                    "Object Name": "Jug",
                    "Title": "Jug",
                    "Artist Display Name": [],
                    "Object Begin Date": "1600",
                    "Medium": "Tin-glazed earthenware",
                    "Classification": [
                        "Ceramics-Faience"
                    ],
                    "Tags": [
                        "Cranes",
                        "Donkeys",
                        "Trees"
                    ]
                },
                "187863": {
                    "Department": "European Sculpture and Decorative Arts",
                    "Object Name": "Bottle",
                    "Title": "Bottle (Refredador)",
                    "Artist Display Name": [],
                    "Object Begin Date": "1500",
                    "Medium": "Tin-glazed and luster-painted earthenware",
                    "Classification": [
                        "Ceramics-Pottery"
                    ],
                    "Tags": [
                        "Coat of Arms"
                    ]
                },...
            }
        },...
    ]
}
\end{lstlisting}
\caption{The excerpt of the raw format of our data, showing an exhibition by the Met Museum, from the year 2000 (\url{https://www.metmuseum.org/exhibitions/listings/2000/spanish-renaissance}), displaying the title and overview text describing the purpose and essence of the exhibition, and some of the mappings to the actual \texttt{object\_ids} and their respective metadata from \citet{metmuseum_openaccess}.}
\label{all_exhibitions_json}
\end{listing}

On the second part of this dataset, we used the Met Museum's Open Access CSV in \citet{metmuseum_openaccess} data, which contains the museum's entire collection index of works of art (a total of \num{484956} works of art) and its associated metadata. The majority of these objects did not show up on any exhibition on the list of exhibitions that we describe above. The metadata consists of many fields (54 in total), but we focus only on ``Department'', ``Object Name'', ``Title'', ``Artist Display Name'', ``Object Begin Date'', ``Medium'', ``Classification'', ``Tags'', together with ``Object ID'' to link to the exhibitions' extract. Note that later on, for the machine learning approaches we ignore ``Object Name'', ``Title'', as these are somehow too arbitrary to be used as meaningful metadata for our modelling. The count of the top strings in each of the fields we consider (``Department'', ``Artist Display Name'', ``Object Begin Date'', ``Medium'', ``Classification'', ``Tags'') is depicted in Table \ref{top_generalised_fields}. We shall call these set of all these fields generalised tags (as opposed to only tags, a name which is already used by the Met Museum as a column name in its dataset of all artworks.) There are a total of $\num{69566}$ generalised tag strings in our full dataset of 236 exhibitions, of which $8591$ are unique generalised tags strings. Below we display a very small subset of this dataset in Table \ref{tab:art_objects} as an example for the reader.

\begin{table}[htb]
\centering
\scriptsize
\begin{tabular}{llr}
\toprule
\textbf{Field} & \textbf{Value} & \textbf{Count} \\
\midrule
Department & Asian Art & 2893 \\
           & Photographs & 1720 \\
           & Drawings and Prints & 1634 \\
           & European Sculpture and Decorative Arts & 1506 \\
           & Modern and Contemporary Art & 725 \\
\midrule
Artist Display Name & Pablo Picasso & 484 \\
                    & Galerie Louise Leiris & 273 \\
                    & Hidalgo Arnera & 165 \\
                    & Aldo and Piero Crommelynck & 126 \\
                    & American Tobacco Company & 121 \\
\midrule
Object Begin Date & 1700 & 617 \\
                  & 1750 & 351 \\
                  & 1800 & 347 \\
                  & 1900 & 343 \\
                  & 1600 & 199 \\
\midrule
Medium & Gelatin silver print & 449 \\
       & Oil on canvas & 277 \\
       & Photomechanical print & 250 \\
       & Albumen silver print from glass negative & 223 \\
       & Etching & 197 \\
\midrule
Classification & Prints & 1776 \\
               & Photographs & 1438 \\
               & Paintings & 1319 \\
               & Drawings & 660 \\
               & Creche & 464 \\
\midrule
Tags & Men & 2340 \\
     & Women & 1382 \\
     & Portraits & 797 \\
     & Flowers & 521 \\
     & Landscapes & 369 \\
\bottomrule
\end{tabular}
\caption{Top 5 values per field or generalised tag count in the Met Museum exhibition data.}
\label{top_generalised_fields}
\end{table}

\begin{table}[htb]
\centering
\footnotesize
\begin{tabular}{l r}
\toprule
\textbf{Field} & \textbf{Non-empty Entries} \\
\midrule
Department & \num{484956} \\
Object Name & \num{484956} \\
Title & \num{484956} \\
Artist Display Name & \num{282513} \\
Object Begin Date & \num{484956} \\
Medium & \num{484950} \\
Classification & \num{406239} \\
Tags & \num{192455} \\
\midrule
\textbf{All Fields Non-empty} & \num{120713} \\
\bottomrule
\end{tabular}
\caption{Non-empty Fields in Metropolitan Museum of Art Dataset. Notice that the fields  ``Object Name'' and ``Title'' are used for information purposes only, and do not enter the machine learning approaches.}
\label{stats_csv}
\end{table}

\begin{table}[htbp]
\centering
\scriptsize
\begin{tabular}{l|l|l|l}
\hline
\textbf{Generalised tags} & \textbf{} & \textbf{} & \textbf{} \\
\hline
Object ID 				& \scriptsize 358770 						& \scriptsize 483355 									& \scriptsize 490021 \\
Department 				& \scriptsize Drawings and Prints 			& \scriptsize Modern and Contemporary Art 				& \scriptsize Modern and Contemporary Art \\
Object Name 			& \scriptsize Print 						& \scriptsize Painting 									& \scriptsize Painting \\
Title 					& \scriptsize Les Deux Baigneuses\ldots 	& \scriptsize Erotic Scene (La Douceur) 				& \scriptsize Still Life with Mandolin and\ldots \\
Artist Display Name 	& \scriptsize Auguste Renoir 				& \scriptsize Pablo Picasso 							& \scriptsize Pablo Picasso \\
Object Begin Date 		& \scriptsize 1895 							& \scriptsize 1903 										& \scriptsize 1924 \\
Medium 					& \scriptsize Etching on wove paper\ldots	& \scriptsize Oil on canvas 							& \scriptsize Oil and sand on canvas \\
Classification 			& \scriptsize Prints 						& \scriptsize Paintings 								& \scriptsize Paintings \\
Tags 					& \scriptsize Female Nudes\textbar{}Bathing & \scriptsize Men\textbar{}Female Nudes 				& \scriptsize Food\textbar{}Mandolins\textbar{}Still Life \\
\hline
\end{tabular}
\caption{\normalsize Excerpt of the Met Museum database of all its \num{484956} artworks. We select only a few metadata fields, which we call generalised tags.
The total number of fields present on the database is 54 in total, but we use only ``Department'',  ``Artist Display Name'', ``Object Begin Date'', ``Medium'', ``Classification'', ``Tags'', together with ``Object ID'' to link to the exhibitions' extract. Above we display ``Object Name'', and ``Title'' for information purposes only, these are not used in the modelling. We truncate some of the values of the metadata fields with ``\ldots'' for presentation only in this table, we use the full strings in our modelling code. For some of the values we have more than one string, e.g., ``Female Nudes\textbar{}Bathing'', which means we have a list of strings, two strings in this particular case. The ``\textbar{}'' or {\em pipe} symbol is used in comma separated files (csv files) as the one given by the Met Museum in its Application Programming Interface (API) \citep{metmuseum_openaccess} to expose its database to external users.}
\label{tab:art_objects}
\end{table}

The dataset is very extensive, and we are not interested in all of those fields, what we shall call hereafter generalised tags. In fact, not all generalised tags are filled for all artworks, which is not surprising, given that some artworks are centuries or millennia old. We summarise in Table \ref{stats_csv} some of these statistics.

Our goal is to map, using machine learning techniques, the title and overview text, to the fields named above or what we call generalised tags, and therefore via those to select artworks from the full set of \num{484956} works of art in the Met Museum.

\section{Methodologies and results}
\begin{figure}[htb]
\centering
\scalebox{0.8}{
\begin{tikzpicture}[x=2cm, y=1cm]

  \tikzstyle{neuron}=[circle, draw=black, minimum size=14pt, inner sep=0pt]
  \tikzstyle{annot} = [text width=4em, text centered]

  \foreach \i in {1,...,4}
    \node[neuron] (I\i) at (0,2.5-\i) {};

  \foreach \i in {1,...,5}
    \node[neuron] (H1\i) at (1.5,3-\i) {};

  \foreach \i in {1,...,6}
    \node[neuron] (H2\i) at (3,3.5-\i) {};

  \foreach \i in {1,...,3}
    \node[neuron] (O\i) at (4.5,2-\i) {};

  \foreach \i in {1,...,4}
    \foreach \j in {1,...,5}
      \draw[->] (I\i) -- (H1\j);
  \foreach \i in {1,...,5}
    \foreach \j in {1,...,6}
      \draw[->] (H1\i) -- (H2\j);
  \foreach \i in {1,...,6}
    \foreach \j in {1,...,3}
      \draw[->] (H2\i) -- (O\j);

  \node[annot, above of=I1, node distance=1.0cm] {Input\\Layer ($x$)};
  \node[annot, above of=H11, node distance=1.0cm] {Hidden\\Layer 1};
  \node[annot, above of=H21, node distance=1.0cm] {Hidden\\Layer 2};
  \node[annot, above of=O1, node distance=1.0cm] {Output\\Layer ($y$)};

  \node[shape=single arrow, draw=black, fill=black,
        minimum height=1cm, minimum width=1.5cm, aspect=2,
        shape border rotate=0] (inArrow) at (-0.5,-0) {}; 
  \node[align=center, rotate=90] at (-1,0) {\textbf{Exhibition title and description}}; 

  \node[shape=single arrow, draw=black, fill=black,
        minimum height=1cm, minimum width=1.5cm, aspect=2,
        shape border rotate=0] (outArrow) at (5.0,0) {}; 
  \node[align=center, rotate=90, text width=7cm] at (5.8,0) {
    \textbf{List of generalised tags and/or\\
            probability of generalised tags and/or\\
            list of artworks}
  };

\end{tikzpicture}
}
\caption{Example of a toy model neural network mapping exhibition data, in this case the title and description (overview text) of the exhibition to the generalised tags or a list of artwork suggestions to assign to the input exhibition. Each node or neuron on the neural network represents a real number, the level of the neuron activity, and the links between the nodes are called weights and represent the influence of that neuron on other neurons.}
\label{fig:conceptual-nn}
\end{figure}
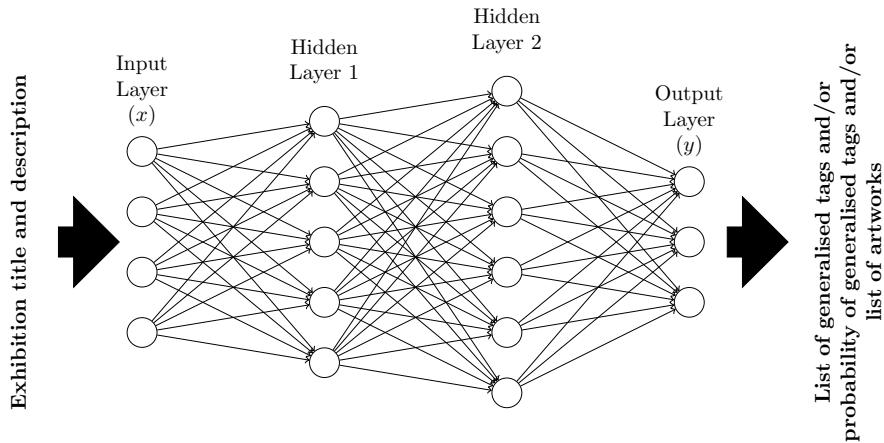

We attempt this mapping between inputs and outputs via four distinct approaches. All our approaches draw on methods developed in the field of Natural Language Processing (NLP) -- see e.g., \citet{young2018recent} for a review and references therein. Schematically, all our approaches can be summarised by Figure \ref{fig:conceptual-nn}, which displays a toy model where we input an exhibition title and description, and output either a list of generalised tags (which is then mapped to the nearest objects that match those tags), or an actual direct list of artworks. Naturally, the neural networks we use are much larger, both in depth and width. Furthermore, some of those neural networks can be much more complex, incorporating other types of connections including loops.

First, we use just the data as it is, without external help, by using its internal statistics. That is to say, we use the probability of finding a word in English within the corpus\footnote{A corpus is a collection of texts that are used for linguistic/statistical analysis.} of exhibition titles and descriptions. In other words, we create a model to represent those words within the language patterns of that corpus. In NLP, this is usually done by using the so-called text vectorisation, which computes word (more accurately, it uses tokens, which can be words, subsets of words or letters) frequency and relative distribution within the corpus \citep{salton1975vector}. Therefore, we create a text vectorisation based on the contents of all the titles concatenated with the overview (description) texts  for all exhibitions taken together. That is to say, we transform the input texts into numeric vectors using the self-contained statistics of the exhibition titles and texts \citep[see e.g.,][on embeddings]{DBLP:journals/corr/abs-1301-3781}. 
Second, we now use external help, in the form of the state-of-the-art OpenAI embedding model \texttt{text-embedding-3-large} that can perform very high-quality text vectorisation. In this sense, we are using the statistical analysis performed by OpenAI on their (much) larger corpus of texts to determine the best (numeric) vectors that represent strings of text. These models also have the advantage of capturing semantic similarity in a metric way \citep[see e.g.,][for details of semantic similarity in NLP]{corley2005measuring}, i.e., they convert textual data into numeric vectors where similar words/sentences/paragraphs have similar vector representation, being near, in the high-dimensional space\footnote{We note here that the dimension of these spaces can be very large, typically of the order of thousands or even tens of thousands of dimensions, making it impossible to clearly visualise these vector representations.} characterising the texts. In our first model, we do not capture semantic similarity, only word frequency, so this second approach, has, in principle, some advantages. 
Third, we use OpenAI embedding models to transform both the input, the title and description of the exhibitions, and the output, for which we use the concatenated strings made of all the metadata associated with all the artworks for each of one the exhibitions. We then can find the nearest embedding vector for an input text and from that find the nearest artworks' metadata embedding vectors for that text, giving us the nearest artworks.
Fourth, and finally, we use the full power and knowledge encapsulated in the OpenAI models, by using the OpenAI Application Programming Interface (API\footnote{An API is a collection or set of rules and specifications that allows to call code or databases remotely without having to store or host the entire infrastructure locally.}) to call OpenAI models to ask for a direct and full list of generalised tags suggested for a given exhibition description (title concatenated with the overview text). We therefore train one of the latest OpenAI models, ``\texttt{gpt-4o-mini}'', using a technique called fine-tuning \citep{devlin2019bert, yosinski2014transferable}. Fine-tuning models is a common approach in the field of neural networks, taking an already pre-trained network and improving it (or fine-tuning it) to a more specific domain
\citep[see e.g.,][for two examples in astrophysics and in finance]{covas2020transfer, covas2023named}. After fine-tuning, we ask the resulting model via the API for lists of generalised tags (without forcing the number of predicted artworks) given an input text (which the fine-tuned model will decided how to represent and embed numerically). 

We note that, for the first, second and third models, we do not take the approach of directly attempting to predict the full list of artworks from the input text, i.e., the text as depicted in Listing \ref{all_exhibitions_json}, or in other words, the number of artworks. This is because, first, it is very difficult to predict a variable number of text strings, i.e., the metadata for the several artworks. Actually, we found that there was no significant correlation between the statistics of the word distribution within the input texts and the number of artworks in each exhibition. Second, because the entire database of artworks is much larger (\num{484956} artworks than  the number of artworks showing up in our exhibition dataset (\num{10470} artworks). Therefore, for the first, second and third models, we impose the number of artworks to be either the number of the artworks of the actual exhibition (for in-sample prediction\footnote{This is what a prediction made using the same data that was used to train the model is called.}) or 16 artworks (for out-of-sample prediction\footnote{Conversely, this is what a prediction made on new data not used during model training is called. These kinds of predictions are made to assess model generalization.}). For the fourth model, the full GPT model, we can let the algorithm decide itself. 

For all approaches, we divide our dataset in Table \ref{exhibitions_table} into a training and validation sets, as it is standard in machine learning approaches. 
We do this in the proportion $80\%$ examples for training (a total of 188 examples) and $20\%$ samples for validation or in-sample testing (a total of 48 examples), taking the total to the above-mentioned global set of 236 exhibitions.\footnote{The validation set is a subset of examples drawn from the same sample -- this means it presumably follows the same probability distribution function -- but is not used for training, just for checking accuracy. On the other hand, test sets can be in-sample (from the same probability distribution) or out-of-sample (from possibly a different distribution), and the latter are used to test the ability of the model to actually generalise.} We train the models on the training set,\footnote{We train and optimise the neural network using the so-called Adam optimiser \citep{kingma2014adam}, which is implemented within Google's TensorFlow deep learning framework \citep{abadi2016tensorflow} in our Python code.} and then test our ability to predict the appropriate metadata (``Department'',``Artist Display Name'', ``Object Begin Date'', ``Medium'', ``Classification'', ``Tags'') and ultimately the artworks for the input text in the validation set, comparing the predicted versus actual metadata or generalised tags and the list of predicted versus actual artworks. All our training optimises to minimise the error difference between the actual output (on the training set) and the predicted output. We also calculate the error difference within the validation set to monitor over-fitting, a common problem with machine learning models \citep{geman1992neural}. Only after training the model, and checking the results for the training and validation sets, we then test out-of-sample on a made-up test set, where we create an input hypothetical exhibition title and description, in the hope that our model can select appropriate artworks for it.


\subsection{Self-contained text vectorisation}
\label{self_contained_text_vectorisation}

Text vectorisation transforms human language in the form of text into a series of numbers, 
of fixed or variable length, that represent that text in a machine-readable format \citep{DBLP:journals/corr/abs-1301-3781}. That way, we can transform the title and the overview text (concatenated together) into a numerical vector -- a series of ordered numbers (and we choose this to be of fixed size to keep the model complexity under control and to have a fixed number of input nodes on our neural network). This will serve as the $x$ input in our machine learning approach, a neural network, as depicted in the conceptual toy model in Figure \ref{fig:conceptual-nn}. 
As implied by the toy model diagram, we therefore also need an output $y$ variable. One approach would be to have the actual artworks and metadata to be a running text output, like what we have on the raw data as depicted in the Listing \ref{all_exhibitions_json} above. However, this would introduce two hard-to-solve challenges: first, the output is of variable size (length), since an exhibition can have any number of associated artworks, and also the artworks can have any number of metadata items associated with them; and second it would not be possible to constrain the neural network to select only artworks from the available \num{484956} works of art in the Met Museum. The first issue could potentially be solved by a so-called Long Short-Term Memory (LSTM) neural network \citep{hochreiter1997long}. These are used, for example, in one human language to another human language translation (or computer programming language to another), and have the advantage of allowing for variable input/output length.\footnote{This ability to achieve variable output length is reached by using a neural network architecture with feedback loops, called a recurrent architecture \citep{elman1990finding}. The neural network outputs a token (or word) at a time, and then that output token is concatenated with the input and fed back into the network until a special end-of-sequence (EOS) token is produced. This allows the model to have a flexible output length.} Nonetheless, even with a LSTM network, and with only a small number of training examples like we have, it would be very difficult to get it to output a consistent format like the one we have for our raw data, with all those fields and fixed format metadata. Furthermore, we really want to constrain our machine learning method to  the range of Met Museum artworks only, and not to a hypothetical range of all world artworks. Therefore, we have devised a simple approach that consists of two steps, first in flattening the string values of the fields (``Department'', ``Artist Display Name'', ``Object Begin Date'', ``Medium'', ``Classification'', ``Tags'') for the training set in the format as depicted in Listing \ref{all_flat_exhibitions_json}, and second, in calculating, from the count of those values (per exhibition), the probability of each value showing in each exhibition, as depicted in the excerpt in the Listing \ref{all_flat_percentages_exhibitions_json}. 

\begin{listing}[htb]
\begin{lstlisting}
{
    "exhibitions": [
        {
	"x": "Title of exhibition is: Sculpture and Decorative Arts of the Spanish Renaissance and the description is: The Metropolitan Museum of Art's small but excellent collection of Spanish polychrome sculpture, including sacred reliefs and freestanding carved figures once housed in the churches of Spain, is displayed in the gallery adjacent to the newly reopened Velez Blanco Patio. The selection, which displays the unique blending of early western European and Islamic stylistic and technical influences, emphasizes the diversity in the material culture of Renaissance Spain after the Catholic reconquest by Ferdinand and Isabella.",
	"y": {
                "Department": [
                    "European Sculpture and Decorative Arts",
                    "European Sculpture and Decorative Arts",
                    "European Sculpture and Decorative Arts",
                    "The American Wing",
                    "European Sculpture and Decorative Arts",
                    "European Sculpture and Decorative Arts",
                    "European Sculpture and Decorative Arts",
                    "European Sculpture and Decorative Arts",
                    "European Sculpture and Decorative Arts",
                    "European Sculpture and Decorative Arts",
                    "European Sculpture and Decorative Arts"
                ],
                "Artist Display Name": [
				
				...,
				
				], ...,
			},
		}, ...,
	], ...,
}
\end{lstlisting}
\caption{Flat version of the dataset for the Met Museum's exhibitions, where $x$ is the title/description of the exhibition and $y$ is the flattened version of the metadata associated with the exhibition. We have therefore eliminated the artworks object ids.}
\label{all_flat_exhibitions_json}
\end{listing}

\begin{listing}[htb]
\begin{lstlisting}
{
    "exhibitions": [
        {
            "x": "Title of exhibition is: Sculpture and Decorative Arts of the Spanish Renaissance and the description is: The Metropolitan Museum of Art's small but excellent collection of Spanish polychrome sculpture, including sacred reliefs and freestanding carved figures once housed in the churches of Spain, is displayed in the gallery adjacent to the newly reopened Velez Blanco Patio. The selection, which displays the unique blending of early western European and Islamic stylistic and technical influences, emphasizes the diversity in the material culture of Renaissance Spain after the Catholic reconquest by Ferdinand and Isabella.",
            "y": {
                "European Sculpture and Decorative Arts": 0.90909091,
                "The American Wing": 0.09090909,
                "Diego de Pesquera": 0.25,
                "Juan Martinez Montanes": 0.25,
                "Juan de Ancheta": 0.25,
                "Diego de Atienza": 0.25,
                "1600": 0.18181818,
                "1500": 0.09090909,
                "1585": 0.27272727,
                "1630": 0.09090909,
                "1567": 0.09090909,
                "1615": 0.09090909,
                "1575": 0.09090909,
                "1646": 0.09090909,
                "Tin-glazed earthenware": 0.09090909,
                "Tin-glazed and luster-painted earthenware": 0.27272727,
                "Silver gilt, enamel": 0.09090909,
                "Wood, painted and gilt": 0.09090909,
                "Wool, silk, metal thread on canvas": 0.18181818,
                "Polychromed wood with gilding": 0.09090909,
                "Wood, polychromed and gilded": 0.09090909,
                "Silver gilt with enamel, cast, chased, and engraved": 0.09090909,
                "Ceramics-Faience": 0.1,
                "Ceramics-Pottery": 0.3,
                "Sculpture": 0.3,
                "Textiles-Embroidered": 0.2,
                "Metalwork-Silver": 0.1
            },
            "z": [
                "187702",
                "187863",
                "196434",
                "197089",
                "199674",
                "210828",
                "210826",
                "201910",
                "202718",
                "205084",
                "197090"
			],
		}, ...,
	], ...,
}
\end{lstlisting}
\caption{Probability flattening of the dataset, where now the $y$ variable represents the percentage of each metadata string occurring within the metadata for all the artworks associated with each particular exhibition, ignoring the metadata type -- that is, we merge all metadata together for each exhibition. This allows a clear mapping between an exhibition title/description and a fixed-size high dimensional probability vector.}
\label{all_flat_percentages_exhibitions_json}
\end{listing}

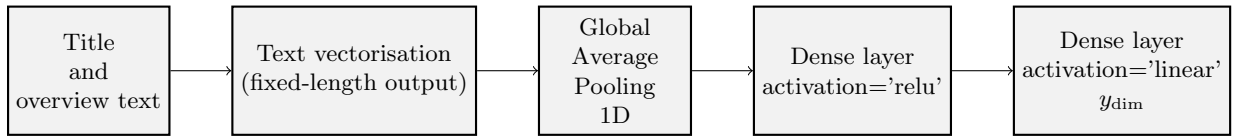
\begin{figure}[htb]
\centering
\small 
\scalebox{0.85}{
\begin{tikzpicture}[
    box/.style={draw, thick, fill=gray!10, minimum width=2.0cm, minimum height=1.7cm, align=center},
    arrow/.style={thick, -{Latex[length=3mm]}},
    node distance=0.8cm and 0.8cm
  ]

  \node[box] (title) {Title\\ and\\ overview text};
  \node[box, right=of title] (vectorisation) {Text vectorisation\\(fixed-length output)};
  \node[box, right=of vectorisation] (gap) {Global\\Average\\Pooling\\1D};
  \node[box, right=of gap] (dense1) {Dense layer\\activation='relu'};
  \node[box, right=of dense1] (dense2) {Dense layer\\activation='linear'\\$y_{\text{dim}}$};

  \draw[->] (title) -- (vectorisation);
  \draw[->] (vectorisation) -- (gap);
  \draw[->] (gap) -- (dense1);
  \draw[->] (dense1) -- (dense2);

\end{tikzpicture}
}
\caption{Neural network approach using self-contained text vectorisation and the (fixed-length) probability of finding a value on the exhibitions metadata fields. From left to right, the neural network takes a variable length plain text, i.e., the title and description of the exhibition, then it creates a vectorisation, which outputs a numeric vector, an embedding of that text. That vector has 256 dimensions, or numbers. 
Then it passes that vector to a one-dimensional average pooling layer, which averages those 256-dimensional vectors to smaller 64-dimensional vectors. That is passed to a so-called hidden layer, a dense layer of 256 nodes that has an activation function (in this case a ReLU function) -- the ReLU (Rectified Linear Unit) function is defined as \( \text{ReLU}(x) = \max(0, x) \), and outputs simply the input if it is positive, and zero otherwise. Finally, a second hidden layer with 8615 nodes or neurons and a linear activation function (in this case identity function) maps the results to the probabilities. In the diagram, $y_{\text{dim}}$ is the number of distinct values of the exhibitions' artworks metadata fields (e.g., [``European Sculpture and Decorative Arts'', ``The American Wing'', ``Diego de Pesquera'', ``1585'', ``Sculpture'', \ldots]). Therefore, $y_{\text{dim}}=8615$. The final outputs are probabilities of each one of those metadata fields (or what we call generalised tags) to show up in the list of artworks associated with that input title and description.   We tested several activation layer functions, which decide the type of output. We also played with the number of hidden layer nodes, and settled on 256 nodes. The network/training parameters were: number of epochs = 2048, with the text vectorisation using \texttt{max\_tokens = 32768}, \texttt{output\_sequence\_length = 256}, \texttt{output\_mode = "int"}, and \texttt{standardize = "lower\_and\_strip\_punctuation"}. For all runs we used \texttt{batch\_size = 16}.}
\label{self_contained_text_vectorisation_neural_network}
\end{figure}


We run and test the model using several combinations made of different parameters and different modes of operation. The neural network we use is depicted, in schematic and compact fashion, in Figure \ref{self_contained_text_vectorisation_neural_network}. The data processing starts left to right: we feed in a title and description, the ``$x$'' in Listing \ref{all_flat_percentages_exhibitions_json}, and that text is cleaned and broken up; transformed into a vector; then after several layers, it outputs probabilities of each generalised tag to show up, and we shall use those later to map the most probable artworks from our database.

As described in the previous section, 
we divide our dataset in Table \ref{exhibitions_table} into a training and a validation sets ($80\%$ examples for training (a total of 188 examples) and $20\%$ samples for validation or in-sample testing -- a total of 48 examples, taking the total to the full set of 236 exhibitions. We train the model as described in the previous section, and monitor the error difference between the predicted probabilities of each generalised tag and the actual probabilities (within the training and validation sets).
We display below in Figure \ref{fig:base_mse_pdf} the loss or mean square error (MSE) for the training of the neural network for the entire training and validation sets.

\begin{figure}[htb]
    \centering
    \begin{overpic}[width=0.75\linewidth]{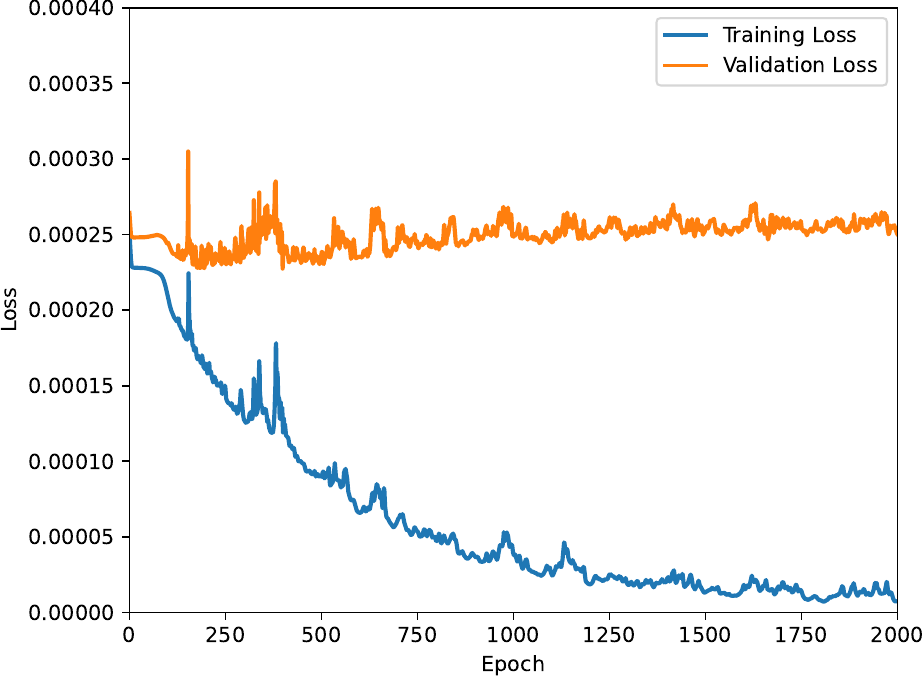}
        \put(54.5,14.4){%
            \includegraphics[width=0.32\linewidth]{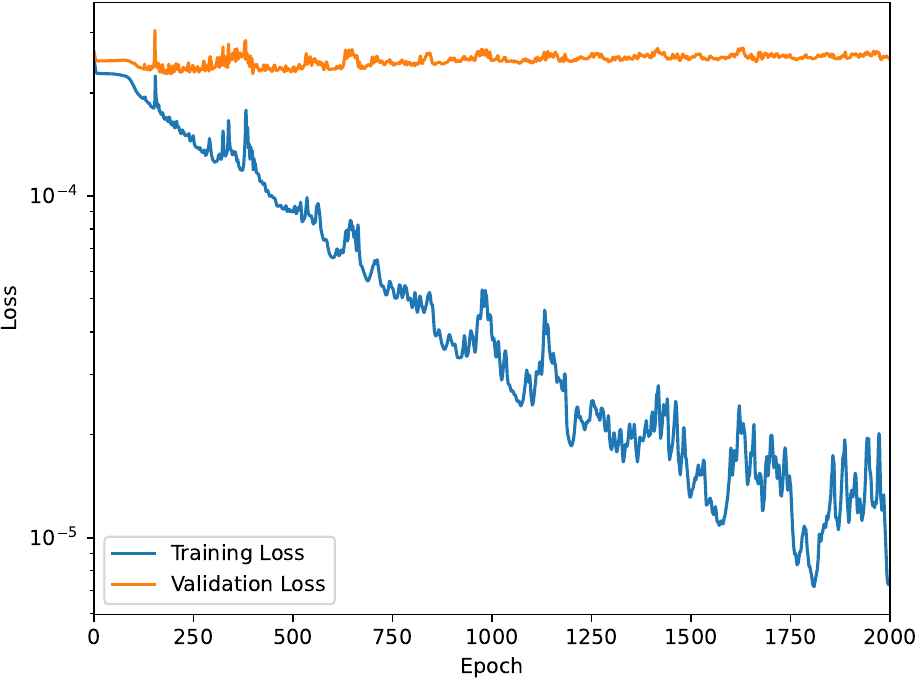}%
        }
    \end{overpic}
    \caption{Mean squared error (MSE) plotted for the training and validation sets of the neural network in Figure~\ref{self_contained_text_vectorisation_neural_network}. The inset shows the same plot with a logarithmic y-axis.}
    \label{fig:base_mse_pdf}
\end{figure}

From analysing Figure \ref{fig:base_mse_pdf} we can conclude two key insights. First, that the models seem to converge as the MSE value decreases with epoch, as training progresses. Second, that it also seems to show that after 150 epochs or so, we start to have a serious case of over-fitting, as we can see the loss on the validation set stops decreasing, so the model is simply memorising the data after that.

\begin{figure}[htb]
    \centering
    \includegraphics[width=0.75\linewidth]{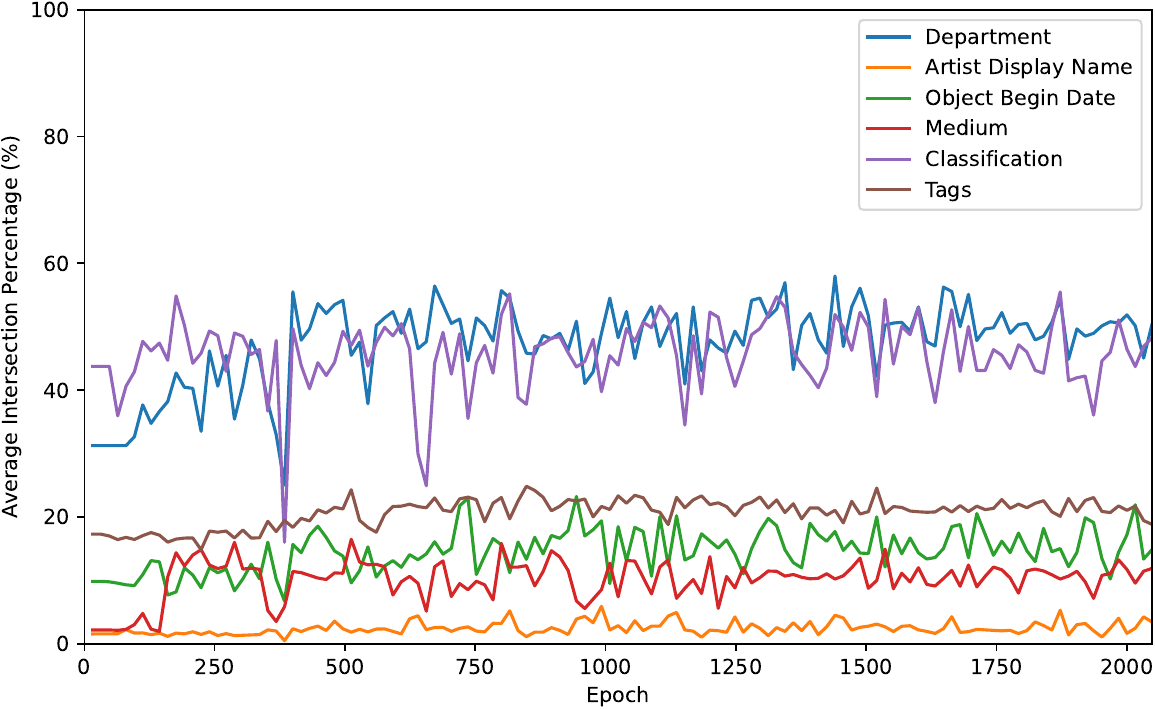}
    \caption{Percentage of generalised tags intersection on the validation set, between actual and prediction by the model.}
    \label{fig:percentage_interception_base_embedding-2048}
\end{figure}

We then attempted to analyse which of the generalised tag sub-groups is our model matching better (or worse). That is, for each exhibition, we look at the actual set of artworks on that exhibition and the predicted one, extract for each the generalised tags per sub-group (``Department'', ``Artist Display Name'', ``Object Begin Date'', ``Medium'', ``Classification'', ``Tags'') and do a comparison, calculating the percentage of matching tags. We then average the percentage intersection or percentage matching across all exhibitions within the validation set (as within the training set this would not make sense, as we can have over-fitting, so we would not learn much by looking at those percentages.). So, e.g., if all the predicted artworks have as Department = ``European Paintings'' and the actual artworks the same, that percentage would be 1, a perfect match. Those results are depicted in Figure \ref{fig:percentage_interception_base_embedding-2048}.

The results show that the model is best at predicting the ``Department'' and ``Classification'' generalised tags, e.g.,
Department = ``Asian Art'', ``Photographs'', ``Drawings and Prints'', ``European Sculpture and Decorative Arts'', ``Modern and Contemporary Art'', etc., and 
Classification = ``Prints'', ``Photographs'', ``Paintings'', ``Drawings'', ``Creche'', etc. The model, however, is quite bad at getting the ``Artist Display Name'', which is not surprising but it still disappointing. So, the model seems to focus more on the style and historical periods or traditions of the art exhibition. It is also quite disappointing that the model cannot achieve more than a $50\%$ matching percentage in any of the generalised tags at all. 

The output is produced as follows. First, we run the model, obtaining an 8615-size numeric vector, like $y_\text{output}=[0.00566933 , 0.01132729 , 0.00091457,   \ldots]$. These are the probabilities of each one of the possible values for the generalised tags within the full dataset of 236 exhibitions we have, e.g.,
[\ldots, 'Cups', 'Curtains', 'Cut Paper', 'Cut and pasted fabrics and papers on paper', \ldots]. Both the output probability vector $p_i$ and the list of all generalised tags $\text{gtags}_i$ have a fixed size of 8615 items. We then calculate a hit score for each one of the rows of the full Met Museum objects or artworks database, which has a total of \num{484956} rows, with columns to be used being ``Department'', ``Artist Display Name'', ``Object Begin Date'', ``Medium'', ``Classification'' and ``Tags''. These columns may have or not one of the generalised tags values $\text{gtags}_i$ above. 
The hit score \( \text{hit}_{j} \) is calculated for row \( j \) in the Met Museum database as:

\begin{equation}
\text{hit}_{j} = \sum_{i=1}^{8615} p_i \cdot \delta_{ij}
\label{hit_function}
\end{equation}

where

\begin{equation}
\delta_{ij} =
\begin{cases}
1, & \text{if } \text{gtags}_i \text{ exists in row } j \\
0, & \text{otherwise}
\end{cases}
\end{equation}

\begin{figure}[htb]
    \centering
    \includegraphics[width=0.75\linewidth]{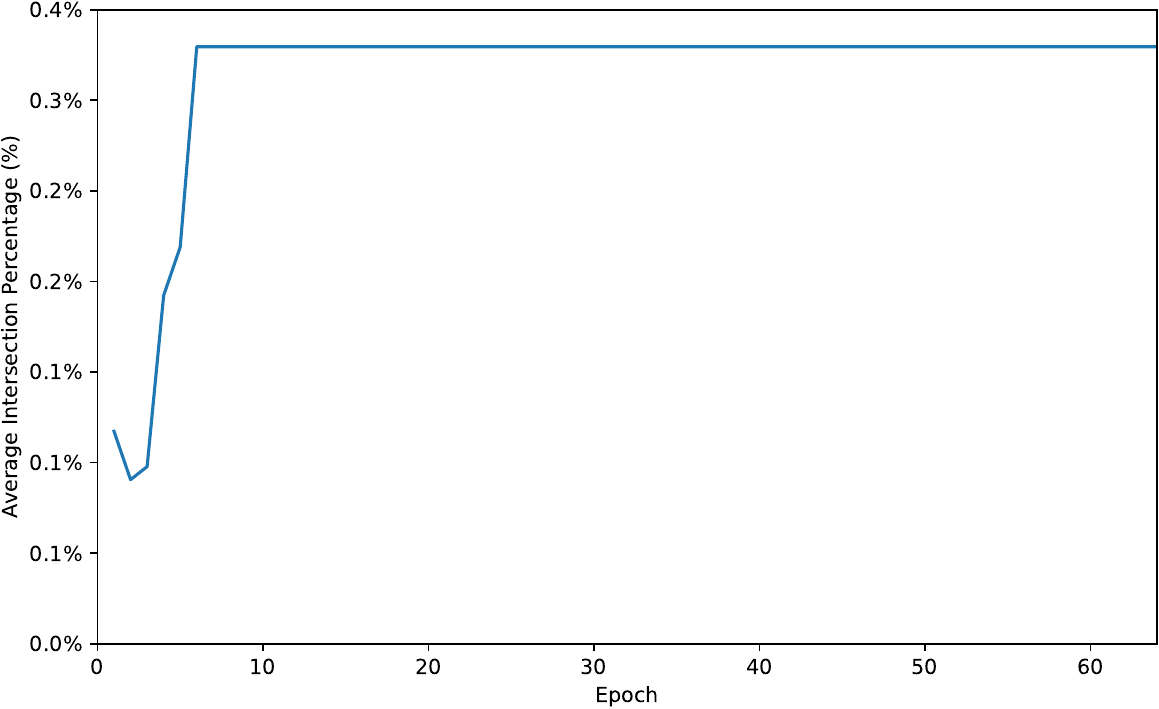}
    \caption{Percentage intersection of predicted artworks that match actual artworks on the validation set.}
    \label{fig:percentage_intersection_artworks_base_openai_64}
\end{figure}

We then sort the database in descending order of $\text{hit}_{j}$, and extract the top 16 artworks that represent the best matches. Given this, we can calculate the percentage of matched artworks, as depicted in Figure \ref{fig:percentage_intersection_artworks_base_openai_64}, is less than $1\%$. Even taken into account that we are trying to match a few artworks across a  \num{400000}+ artworks database, this is still very small. Taken together, the over-fitting, the low matching percentage of the generalised tags and the low matching of artworks seem to indicate the model is weak. We shall show that it is indeed very weak, next, when we test the model out-of-sample.

We tested the model out-of-sample and input the following title and description for a hypothetical exhibition. For this case, we used all the 236 exhibitions as training set, since we are doing out-of-sample prediction, there is no need to restrict ourselves by splitting the full dataset into a training and validation set, it would be wasteful. Notice also that the choice of the description was done on purpose since the word ``impressionist'' only shows up twice in a single exhibition, the Met Museum's ``Neo-Impressionism''\footnote{This exhibition website page can be seen in \url{https://www.metmuseum.org/exhibitions/listings/2001/paul-signac-circle}.} exhibition of 2001, with 7 paintings, and the sentence ``still life'' shows up 50 times within the generalised tags, well below the count for other generalised tags as we can see from Table \ref{top_generalised_fields}. So, we are trying to make it difficult for the machine learning model to predict what generalised tags, and therefore what artworks to choose for our test input exhibition. 

\begin{ttquotesmall}
Title of exhibition is: The First Impressionists in Paris and the description is: the paintings of still life from the first Impressionists who created the modern art.
\end{ttquotesmall}

 The results, using the full training set of 236 exhibitions with a training of 2048 epochs were as follows (results were as bad at 32 and 64 and 150 epochs too, before it starts to do over-fitting):

\begin{table}[htb]
\centering
\tiny
\begin{tabular}{l|l|l|c|c|c}
\hline
\makecell{\textbf{Object}\\ \textbf{ID}} & \textbf{Department} & \makecell{\textbf{Artist}\\ \textbf{Display Name}} & \makecell{\textbf{Object}\\ \textbf{Begin} \\ \textbf{Date}} & \textbf{Medium} & \textbf{Classification} \\
\hline
38056 & Asian Art & [] & 1630 & Ink, opaque watercolor, and gold on paper & [Paintings] \\
64533 & Asian Art & [] & 1760 & Ink, opaque watercolor, and gold on paper & [Paintings] \\
56802 & Asian Art & [Torii Kiyonaga] & 1742 & Woodblock print; ink and color on paper & [Prints] \\
67568 & Asian Art & [] & 1800 & Silk & [Textiles-Woven] \\
37883 & Asian Art & [] & 1620 & Ink and opaque watercolor on paper & [Paintings] \\
71999 & Asian Art & [] & 1700 & Silk / Compound weave & [Textiles-Woven] \\
37812 & Asian Art & [] & 1510 & Ink and opaque watercolor on paper & [Paintings] \\
37897 & Asian Art & [] & 1640 & Ink and opaque watercolor on paper & [Paintings] \\
37920 & Asian Art & [] & 1734 & Ink and opaque watercolor on paper & [Paintings] \\
50939 & Asian Art & [] & 1600 & Ink, opaque watercolor, and gold on paper & [Paintings] \\
51680 & Asian Art & [Unidentified artist] & 1800 & Album leaf; ink and color on silk & [Paintings] \\
671016 & Asian Art & [] & 1500 & Hanging scroll; ink, color and gold on silk & [Paintings] \\
37174 & Asian Art & [Torii Kiyonaga] & 1778 & Woodblock print; ink and color on paper & [Prints] \\
45034 & Asian Art & [Katsushika Hokusai] & 1760 & Woodblock print; ink and color on paper & [Prints] \\
55258 & Asian Art & [Utagawa (Gountei) Sadahide] & 1866 & Triptych of woodblock prints; ink and color on paper & [Prints] \\
59202 & Asian Art & [] & 1800 & Wood & [Netsuke] \\
\hline
\end{tabular}
\label{first_model_output}
\end{table}

\begin{figure}[!htb]
    \centering
    \includegraphics[width=0.75\linewidth]{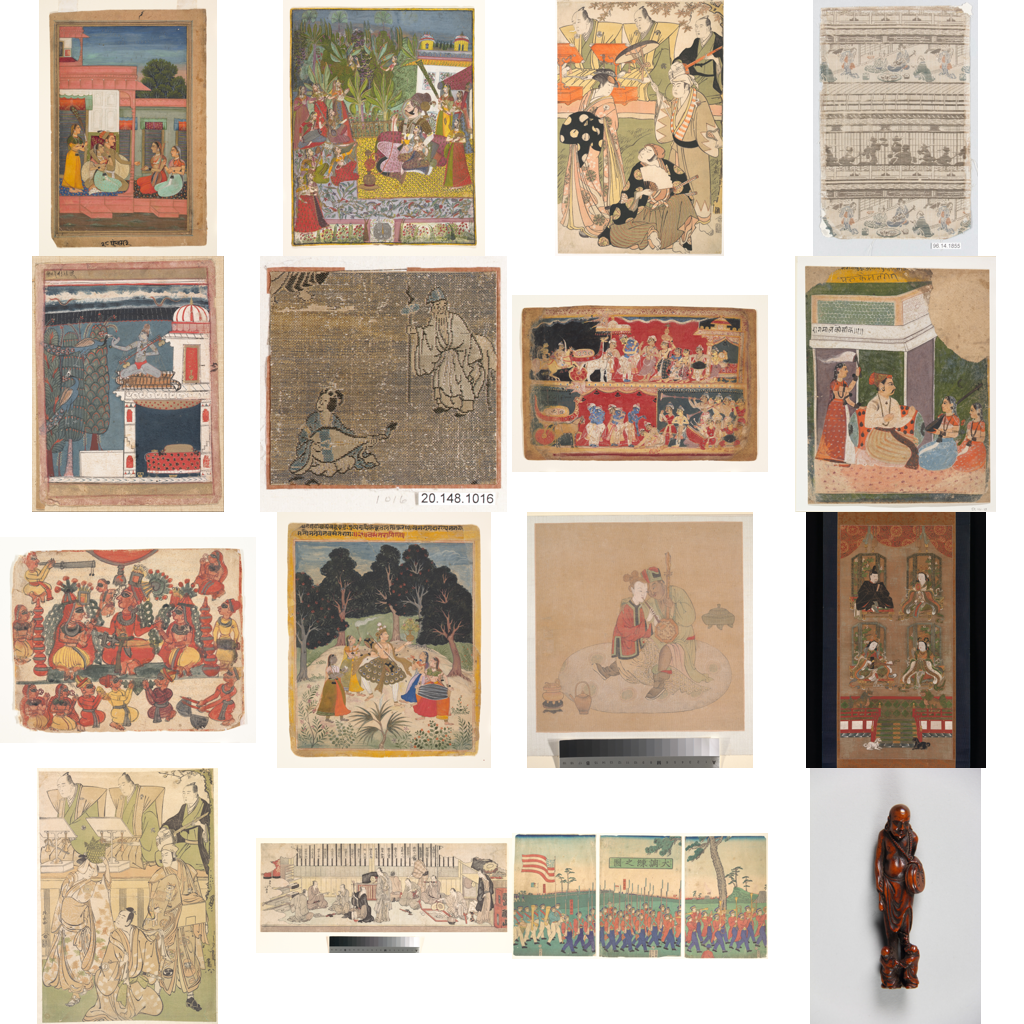}
    \caption{Marble composite of images related to the artworks from the Met Museum corresponding to the run with epochs=2048 for the model training with the entire dataset, using as input self-contained embeddings and output per exhibition metadata occurrence frequencies. Notice we only display the images of the artworks that are on the public domain, as provided by the Met Museum website and its API \citep{metmuseum_openaccess}.}
	\label{marble_base_embedding_epochs=2048_filter_w_tags_236_training}
\end{figure}

The results, the artwork choices, which can be seen in the output above and in the marble composite image in Figure \ref{marble_base_embedding_epochs=2048_filter_w_tags_236_training} show that the results are very poor. All the works are from ``Asian Art'' rather than ``European Paintings'', and the date span is totally wrong, 
1500--1866. This led us to attempt to either improve or choose another machine learning model for our task. Nonetheless, we decided to keep these results for comparison and to have a base (and basic) model. Our first improvement is the subject of the next section.

\subsection{OpenAI-based embedding vectorisation}
\label{openai_embedding_vectorisation}

In this second approach, we use some external help in the form of the state-of-the-art OpenAI embedding model  
\texttt{text\allowbreak-embedding\allowbreak-3\allowbreak-large} that can perform very high-quality 
text vectorisation \citep{openai2024embedding}. So, instead of relying exclusively on the text input (a total of \num{26388} words) and its statistics to create numeric vectors that hopefully will show some syntactic and semantic similarity and statistical analysis that can be used, we shall rely on the much larger training corpus of the OpenAI models, which are known to produce numeric vectors that encode a high-level of syntactic and semantic representation (see e.g., \citet{keraghel2024beyond, Giglietto_2024, petukhova2025text}). We apply this to the input text, the exhibition's title and description (overview text) concatenation, replace the encoding that was done fully internally before,
while keeping the rest of the neural network unchanged, as described in the previous methodology as depicted in Figure \ref{self_contained_text_vectorisation_neural_network}. The resulting network can be seen in Figure \ref{openai_embedding_vectorisation_neural_network} below.

\begin{figure}[htb]
\centering
\small
\scalebox{0.9}{
\begin{tikzpicture}[
    box/.style={draw, thick, fill=gray!10, minimum width=2.0cm, minimum height=1.7cm, align=center},
    arrow/.style={thick, -{Latex[length=3mm]}},
    node distance=0.8cm and 0.8cm
  ]

  \node[box] (title) {Title\\ and\\ overview text};
  \node[box, right=of title] (vectorisation) {OpenAI\\\texttt{text-embedding-3-large}\\(fixed-length output)};
  \node[box, right=of vectorisation] (dense1) {Dense layer\\activation='relu'};
  \node[box, right=of dense1] (dense2) {Dense layer\\activation='linear'\\$y_{\text{dim}}$};

  \draw[->] (title) -- (vectorisation);
  \draw[->] (vectorisation) -- (dense1);
  \draw[->] (dense1) -- (dense2);

\end{tikzpicture}
}
\caption{Neural network approach using OpenAI embedding model \texttt{text-embedding-3-large} and the (fixed-length) probability of finding a value on the exhibitions metadata fields.}
\label{openai_embedding_vectorisation_neural_network}
\end{figure}
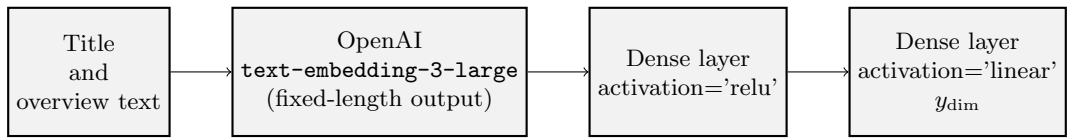

One of the advantages of using the OpenAI embedding models to create numeric vectors from the input text is that these models use the so-called Transformer architecture \citep{vaswani2017attention}. These models take  into account not only the word or token frequency, but also the direction or order of the text (so ``\ldots will many\ldots '' is encoded differently from ``\ldots many will\ldots'') and contextual information such as recognising and recording that the ``he'' in ``\ldots the direction of Picasso's work changed, and he moved away from Cubist approaches\ldots'' is related to ``Picasso''. These models have semantic encoding, recording not just the probability distribution function of words but also the meaning, context and direction and far-away relationships. Furthermore, these models have been trained in a vast corpus of textual data, and therefore have better vector similarity, among other advantages, for semantically similar word sequences, particularly in English language.


\begin{figure}[htb]
    \centering
    \begin{overpic}[width=0.75\linewidth]{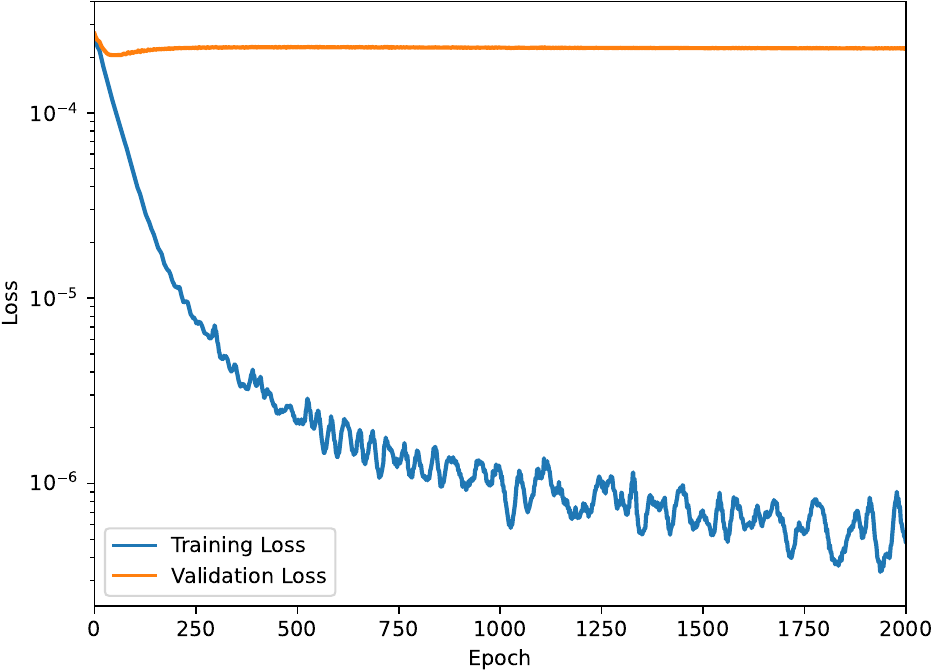}
        \put(40.4,24.5){%
            \includegraphics[width=0.41\linewidth]{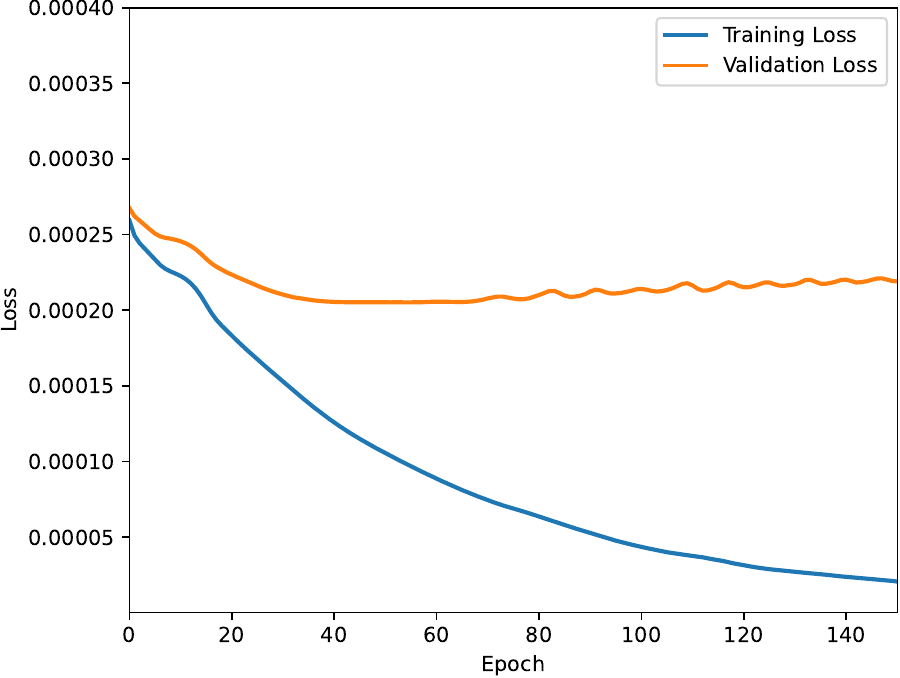}%
        }
    \end{overpic}
    \caption{Mean squared error (MSE) plotted for the training and validation sets for the neural network in Figure \ref{openai_embedding_vectorisation_neural_network}, with an inset zoomed region.}
    \label{fig:openai_embedding_mse_pdf}
\end{figure}

As before, the first thing to do is to train the model, we use the same training and validation sets as before ($80\%$ of examples for training -- a total of 188 examples, and $20\%$ samples for validation or in-sample testing -- a total of 48 examples). We optimise the model using the same numerical approach as before (Adam optimisation). We measure the same mean squared error (MSE) as before, i.e., the sum of squares of the differences in predicted versus actual probabilities of occurrence of each one of the generalised tags (we recall that there are $8591$ unique number of those). The MSE values for the training and validation sets as a function of the number of training epochs is depicted in Figure \ref{fig:openai_embedding_mse_pdf}.

\begin{figure}[htb]
    \centering
    \includegraphics[width=0.75\linewidth]{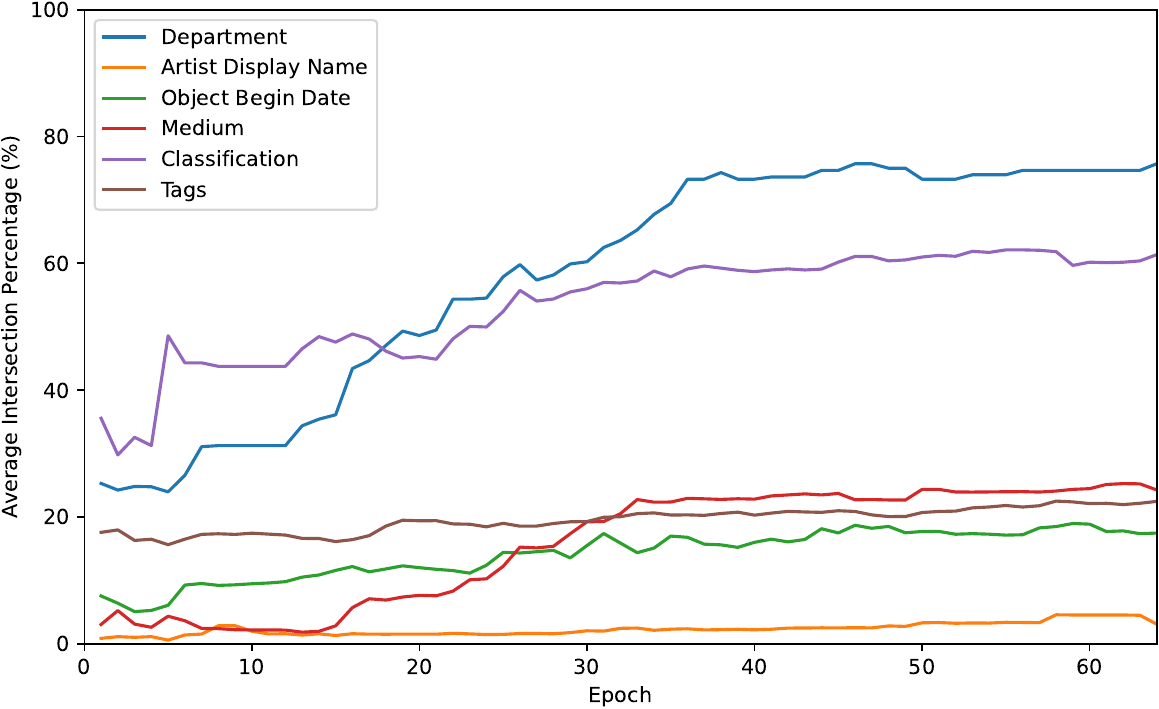}
    \caption{Percentage of generalised tags intersection on the validation set, between actual and prediction by the model.}
    \label{fig:percentage_interception_openai_embedding}
\end{figure}

\begin{figure}[htb]
    \centering
    \includegraphics[width=0.75\linewidth]{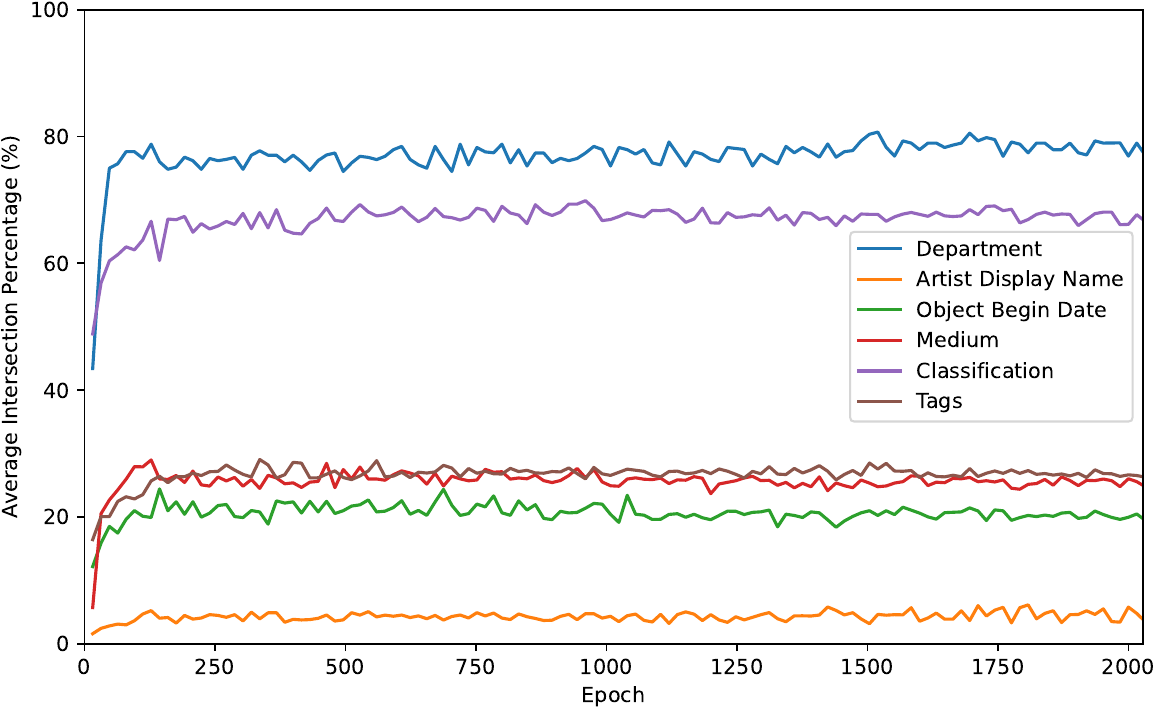}
    \caption{Percentage of generalised tags intersection on the validation set, between actual and prediction by the model.}
    \label{fig:percentage_interception_openai_embedding-2048}
\end{figure}

The results seem to show a slightly better convergence on both the training and validation sets, when one compares with the MSE of the previous model  in Figure \ref{fig:base_mse_pdf}. The MSE in Figure \ref{fig:openai_embedding_mse_pdf} seems to show we reach convergence around 50 training epochs or so. We can also see, as depicted in Figures \ref{fig:percentage_interception_openai_embedding} and \ref{fig:percentage_interception_openai_embedding-2048}, that as the training progresses, the percentage of tags that it gets right increases. Here the results seem much of an improvement with respect to the previous model, if we compare against the results in Figure \ref{fig:percentage_interception_base_embedding-2048}.

\begin{figure}[htb]
    \centering
    \includegraphics[width=0.75\linewidth]{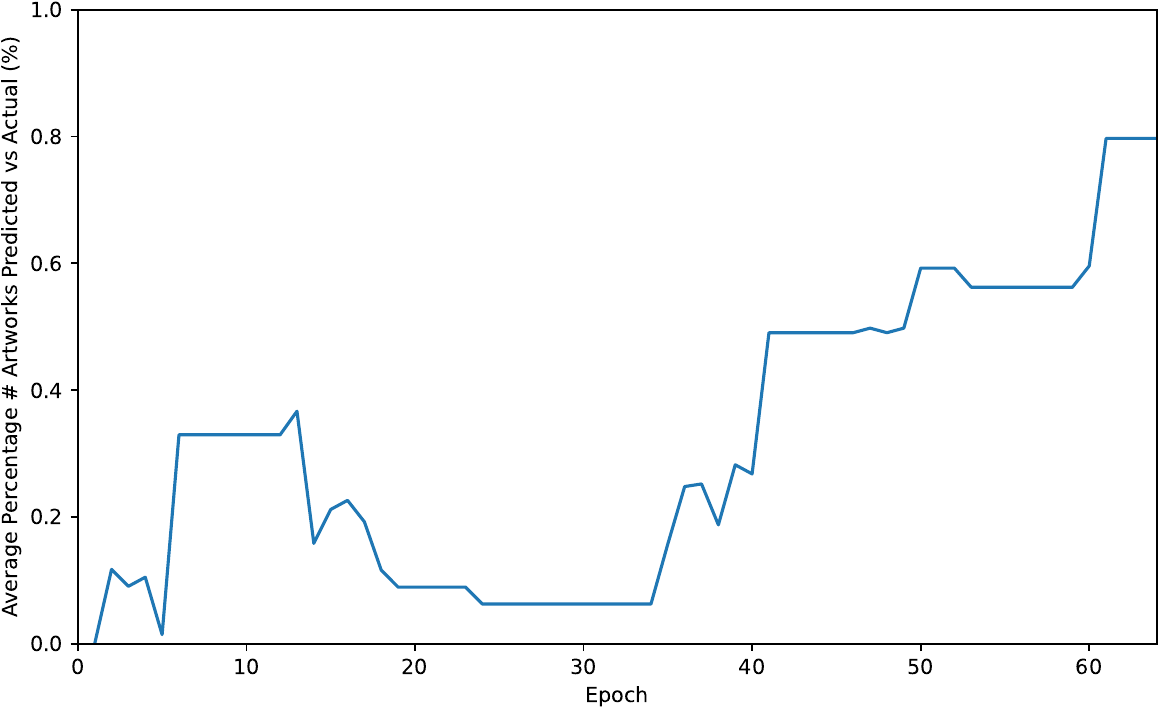}
    \caption{Average percentage of predicted artworks that match actual artworks on the validation set. Note that the percentage
overlap based on random choice is around  $0.00907\%$.}
    \label{fig:percentage_intersection_artworks_openai_embedding_64}
\end{figure}

The results show that, at 64 epochs, the percentage of intersection for each of the generalised tag sub-groups is higher for this model with OpenAI embedding than for the one with self-contained embeddings (previous section). Furthermore, we can see the results are stable for longer runs, as can be seen in the run at 2048 epochs, in Figure \ref{fig:percentage_interception_openai_embedding-2048}. In fact, for ``Department'' and ``Classification'', the matching is around $70\%-80\%$, which is encouraging. We also analyse the average percentage of intersection of predicted artworks per exhibition versus actual artworks within the validation set, and this is depicted in Figure \ref{fig:percentage_intersection_artworks_openai_embedding_64}.

The results seem to show also an improvement with respect to the first model, the base model with self-contained embeddings (depicted in Figure \ref{fig:percentage_intersection_artworks_base_openai_64}). The results jump from around $0.35\%$ to around $0.8\%$. In fact, when we increase the number of epochs to 2048, we get a convergence and saturation of this intersection value at around $1.34\%$. Although this value seems small, one has to think carefully how much larger this value is with respect to the value one obtains if we select randomly $k$ artworks from the full $N=\num{484956}$ works of art dataset. 
If we draw randomly $k=44$ ($44.36$ is the average number of artworks across all our 236 exhibitions) artworks from the full dataset, the average count of artworks expected overlap or intersection is $k \times \frac{k}{N}=\frac{k^2}{N}= \frac{44^2}{484956}=\approx 0.00399$. 
So, the percentage overlap (count of overlap divided by the number of artworks in each exhibition) can be approximated by $k \times \frac{k}{k \times N}=\frac{k}{N}= \frac{44}{484956}=\approx 9.072988 \times 10^{-5} \approx 0.00907\%$. So, for our current model we can get a accuracy 150 times or so better than random choice (even for our basic self-contained model we get an accuracy 70 better than random choice). The question now is, do these better results reflect themselves on better out-of-sample predictions, where these models should really count?







We test this by again running our model out-of-sample, using as training set all 236 example exhibitions we have, and inputting the following title and description for a hypothetical exhibition.

\begin{ttquotesmall}
Title of exhibition is: The First Impressionists in Paris and the description is: the paintings of still life from the first Impressionists who created the modern art.
\end{ttquotesmall}


As we have done before, we run the model to obtain a fixed-size output probability vector $y_\text{output}$, again of length 8615, where each element $p_i$ corresponds to the probability of each of the generalised tags in $\text{gtags}_i$. This is for all generalised tags observed across all 236 exhibitions. We then compute the hit score as before, given by the formula \eqref{hit_function}. Finally, we sort all entries by $\text{hit}_j$ in descending order and select the top 16 artworks with the highest hit scores. The output was, using this time 128 epochs for training:

\begin{table}[htb]
\centering
\tiny
\begin{tabular}{l|l|l|c|c|c}
\hline
\makecell{\textbf{Object}\\ \textbf{ID}}  & \textbf{Department} & \makecell{\textbf{Artist}\\ \textbf{Display Name}} & \makecell{\textbf{Object}\\ \textbf{Begin Date}}  & \textbf{Medium} & \textbf{Classification} \\
\hline
436526 & European Paintings & [Vincent van Gogh] & 1890 & Oil on canvas & [Paintings] \\
436525 & European Paintings & [Vincent van Gogh] & 1890 & Oil on canvas & [Paintings] \\
436528 & European Paintings & [Vincent van Gogh] & 1890 & Oil on canvas & [Paintings] \\
437480 & European Paintings & [Hubert Robert] & 1777 & Oil on canvas & [Paintings] \\
437984 & European Paintings & [Vincent van Gogh] & 1889 & Oil on canvas & [Paintings] \\
437654 & European Paintings & [Georges Seurat] & 1887 & Oil on canvas & [Paintings] \\
437658 & European Paintings & [Georges Seurat] & 1884 & Oil on canvas & [Paintings] \\
436534 & European Paintings & [Vincent van Gogh] & 1890 & Oil on canvas & [Paintings] \\
435849 & European Paintings & [Carolus-Duran (Charles-Auguste-Emile Durant)] & 1890 & Oil on canvas & [Paintings] \\
435868 & European Paintings & [Paul Cézanne] & 1890 & Oil on canvas & [Paintings] \\
436923 & European Paintings & [Maximilien Luce] & 1890 & Oil on canvas & [Paintings] \\
436529 & European Paintings & [Vincent van Gogh] & 1888 & Oil on canvas & [Paintings] \\
436531 & European Paintings & [Vincent van Gogh] & 1885 & Oil on canvas & [Paintings] \\
436536 & European Paintings & [Vincent van Gogh] & 1889 & Oil on canvas & [Paintings] \\
438722 & European Paintings & [Vincent van Gogh] & 1885 & Oil on canvas & [Paintings] \\
437378 & European Paintings & [Odilon Redon] & 1890 & Oil on canvas & [Paintings] \\
\hline
\end{tabular}
\end{table}

\begin{figure}[htb]
    \centering
    \includegraphics[width=0.75\linewidth]{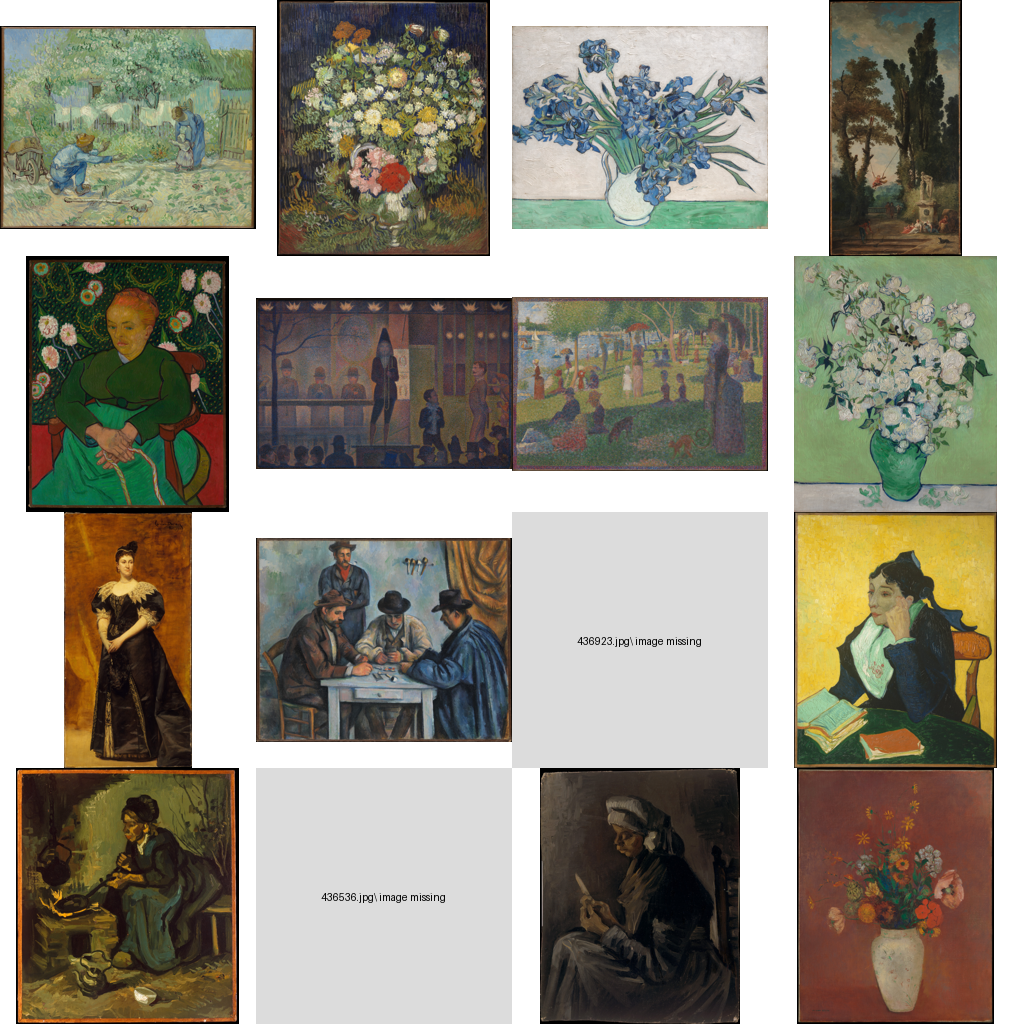}
    \caption{Marble composite of images from the Met Museum corresponding to the run with epochs=128 for the model training with the entire dataset, using as input OpenAI embeddings and output per exhibition metadata occurrence frequencies. Notice we only display the images of the artworks that are on the public domain, as provided by the Met Museum website and its API \citep{metmuseum_openaccess}. Non-available images are marked by their object id JPEG name.}
	\label{marble_openai_embedding_epochs=128_filter_w_tags_236_training}
\end{figure}

Also, the output was, using the longer 2048 epochs training:

\begin{table}[htb]
\centering
\tiny
\begin{tabular}{l|l|l|c|c|c}
\hline
\makecell{\textbf{Object}\\ \textbf{ID}}  & \textbf{Department} & \makecell{\textbf{Artist}\\ \textbf{Display Name}} & \makecell{\textbf{Object}\\ \textbf{Begin Date}} & \textbf{Medium} & \textbf{Classification} \\
\hline
436525 & European Paintings & [Vincent van Gogh] & 1890 & Oil on canvas & [Paintings] \\
436528 & European Paintings & [Vincent van Gogh] & 1890 & Oil on canvas & [Paintings] \\
437378 & European Paintings & [Odilon Redon] & 1890 & Oil on canvas & [Paintings] \\
436526 & European Paintings & [Vincent van Gogh] & 1890 & Oil on canvas & [Paintings] \\
436534 & European Paintings & [Vincent van Gogh] & 1890 & Oil on canvas & [Paintings] \\
436530 & European Paintings & [Vincent van Gogh] & 1888 & Oil on canvas & [Paintings] \\
437984 & European Paintings & [Vincent van Gogh] & 1889 & Oil on canvas & [Paintings] \\
435849 & European Paintings & [Carolus-Duran (Charles-Auguste-Emile Durant)] & 1890 & Oil on canvas & [Paintings] \\
435868 & European Paintings & [Paul Cézanne] & 1890 & Oil on canvas & [Paintings] \\
435880 & European Paintings & [Paul Cézanne] & 1890 & Oil on canvas & [Paintings] \\
436923 & European Paintings & [Maximilien Luce] & 1890 & Oil on canvas & [Paintings] \\
847023 & European Paintings & [Frederic, Lord Leighton] & 1890 & Oil on canvas & [Paintings] \\
437654 & European Paintings & [Georges Seurat] & 1887 & Oil on canvas & [Paintings] \\
437655 & European Paintings & [Georges Seurat] & 1881 & Oil on canvas & [Paintings] \\
437658 & European Paintings & [Georges Seurat] & 1884 & Oil on canvas & [Paintings] \\
438015 & European Paintings & [Georges Seurat] & 1886 & Oil on canvas & [Paintings] \\
\hline
\end{tabular}
\end{table}

\begin{figure}[htb]
    \centering
    \includegraphics[width=0.75\linewidth]{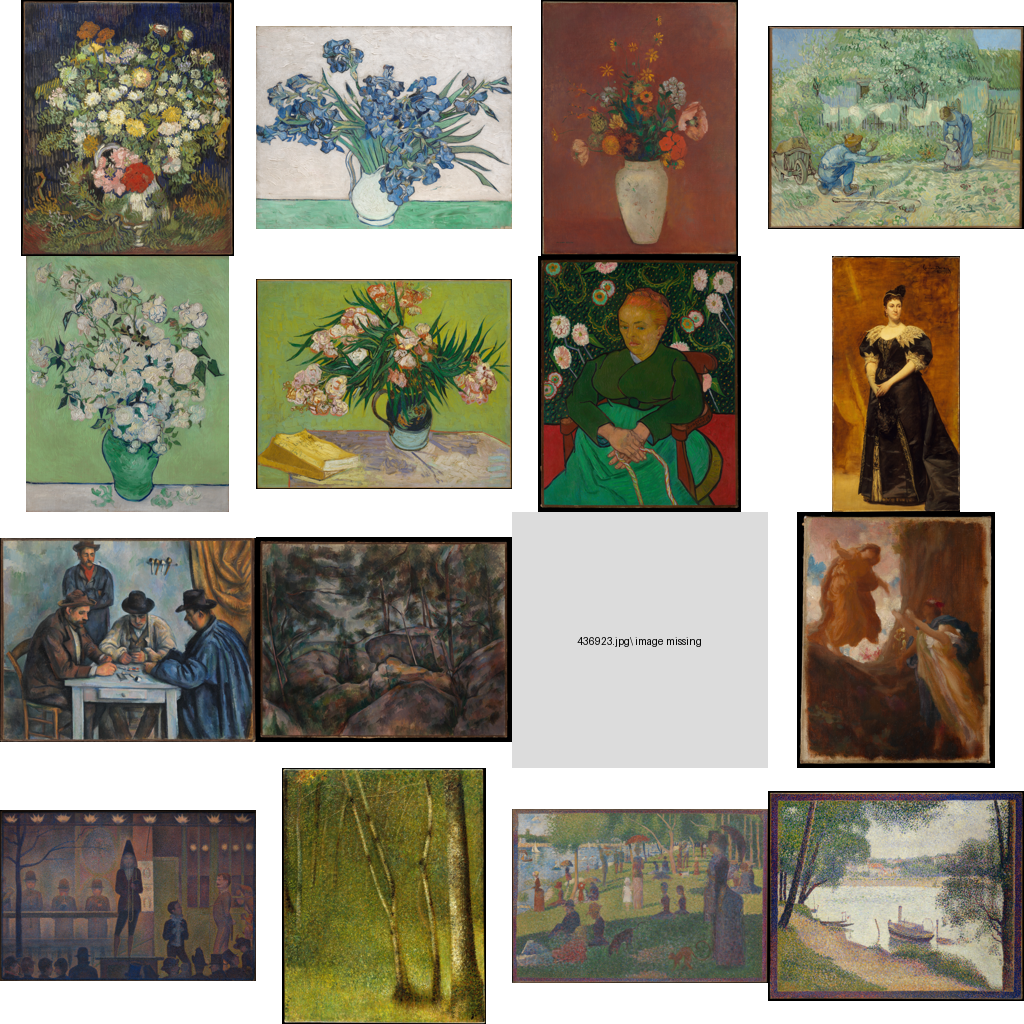}
    \caption{Marble composite of images from the Met Museum corresponding to the run with epochs=2048 for the model training with the entire dataset, using as input OpenAI embeddings and output per exhibition metadata occurrence frequencies. Notice we only display the images of the artworks that are on the public domain, as provided by the Met Museum website and its API \citep{metmuseum_openaccess}. Non-available images are marked by their object id JPEG name.}
	\label{marble_openai_embedding_epochs=2048_filter_w_tags_236_training}
\end{figure}

These results, in particular the second one, with a longer training (even if it over-fits in the training set), are more in line with our expectations that our model is now working better as intended. We can also see that from the marble composite
depicted in Figure \ref{marble_openai_embedding_epochs=2048_filter_w_tags_236_training}. The first element that seems to improve was that we now have a much tighter artwork creation interval, all artworks are within the  period 1777--1890, a span of 113 years for the run with 128 epochs and 1881--1890, a span of only 9 years, for the longer run with 2048 epochs. Impressionism is considered to be a period of art history around 1860--1886 \citep{herbert1988impressionism, Gombrich2023-gh}, although this is not exactly fixed, obviously. Nonetheless, our results are close, even if our system seems to have an inclination to choose a lot of Post-Impressionists and Neo-Impressionists rather than actual from the first Impressionists, as we asked it to do by using as input ``\ldots
First Impressionists in Paris\ldots''. The second element that seems to also improve, is that now we have all artworks  ``Department'' = ``European Paintings'' as   opposed to ``Asian Art'', as it should be, and ``Classification'' as ``Paintings'', not as ``Prints'' or ``Textiles-Woven''. The model also seems to make an effort to find instances of still life, as can be seen by the 4 matches in Figure \ref{marble_openai_embedding_epochs=128_filter_w_tags_236_training} and the
5 matches in Figure \ref{marble_openai_embedding_epochs=2048_filter_w_tags_236_training} -- not only that is a good percentage of matches, but increases with increasing number of training epochs, as it should.\footnote{We note that we can examine freely all the images of all the artworks, even if we cannot display them in this article, by going to the URL \url{https://www.metmuseum.org/art/collection/search/xxxxxx}, where \texttt{xxxxxx} is the \texttt{object\_id}. This way we can decide, by visual inspection, which one are instances of still life or not.}

The results are also interesting in the sense that only $5$ out of the predicted $16$ artworks, $31.25\%$ of them, are present in our full dataset of 236 exhibitions, meaning that the model seems to be able to select artworks outside the training set, and select from the full dataset of $\num{400000}+$ artworks in the Met Museum. This was one of the objectives of this work, to be able to generalise outside the given input exhibition dataset. Therefore, we seem to be going in the right direction by using more advanced embeddings. In the next section, we go one step further in this direction.

\subsection{Direct neural network embedding output to embedding metadata of artworks}
\label{openai_embedding_embedding}



Given that the OpenAI embeddings improved the accuracy of our model, we decided to use it not just for the input but for both the inputs and outputs, i.e., the text/descriptions of the exhibitions and the corresponding concatenated generalised tags for all the artworks in each exhibition.\footnote{We note that there are several ways to encode the output target values, the generalised tags. One alternative would be to encode each exhibition against each  of the generalised tags present in each of the subgroups, or one-to-one even, in effect creating and unfolding more training examples. We decided not to complicate this model too much and simply try to map the encoding or embedding of the title description against the encoding or embedding or a single long string corresponding to all the concatenated generalised tags for all the artworks in each exhibition in the order they show up on the exhibition webpage and the Met Museum database.} We also create the embeddings for all the Met Museum's \num{484956} works of art, for their corresponding generalised tags or metadata and store those on a vector database \citep{andoni2008near}. The reasons we use a vector database are: first, the embeddings data structure becomes too large for our computational resources/capacity available; second, it becomes too slow to search for the best exact nearest neighbours in an array in memory/disk; and third, that these vector databases are optimal for storing embeddings and provide already built-in mechanisms for nearest neighbour search \citep{iscen2017memory}, which will be the approach we take in order to find the artworks (or metadata embeddings) corresponding to the input texts (or title/description embeddings). We then train the model and when that trained model is passed an input text (either from the training or validation sets or from a new out-of-sample test set), it creates an output embedding vector, for which we can then find the nearest neighbours on the vector database of all Met Museum's artworks embeddings. Those will be the selected artworks for that corresponding exhibition. 
We use the Facebook AI Similarity Search (FAISS) library as described in \citet{douze2024faiss}, in particular we use the Inverted File with Flat (IndexIVFFlat) vectors model, since the exact match approach could not fit the memory available on our modest computational resources.
Below in Figure \ref{openai_embedding_to_embedding_vectorisation_neural_network} we can find a schematic of the neural network for this kind of modelling.

\begin{figure}[htb]
\centering
\small
\scalebox{0.6}{
\begin{tikzpicture}[
    box/.style={draw, thick, fill=gray!10, minimum width=2.0cm, minimum height=1.7cm, align=center},
    arrow/.style={thick, -{Latex[length=3mm]}},
    node distance=0.8cm and 0.8cm
  ]

  \node[box] (title) {Title\\ and\\ overview text};
  \node[box, right=of title] (vectorisation) {OpenAI\\\texttt{text-embedding-3-large}\\(fixed-length output)};
  \node[box, right=of vectorisation] (dense1) {Dense layer\\activation='relu'};
  \node[box, right=of dense1] (dense2) {Dense layer\\activation='linear'\\$y_{\text{dim}}=3072$};
  \node[box, right=of dense2] (vectorisation2) {OpenAI\\\texttt{text-embedding-3-large}\\(fixed-length output)};
  \node[box, right=of vectorisation2] (metadata) {Concatenated\\artworks's\\metadata};

  \draw[->] (title) -- (vectorisation);
  \draw[->] (vectorisation) -- (dense1);
  \draw[->] (dense1) -- (dense2);
  \draw[->] (vectorisation2) -- (dense2);
  \draw[->] (metadata) -- (vectorisation2);

\end{tikzpicture}
}
\caption{Neural network approach using the OpenAI embedding model \texttt{text-embedding-3-large} for both the inputs: the title/descriptions of the exhibitions and the outputs, the concatenated flattened list of generalised tags for the exhibitions.}
\label{openai_embedding_to_embedding_vectorisation_neural_network}
\end{figure}
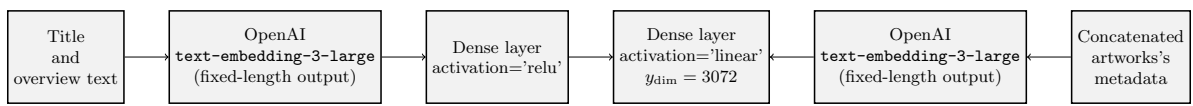

\begin{figure}[htb]
    \centering
    \includegraphics[width=0.75\linewidth]{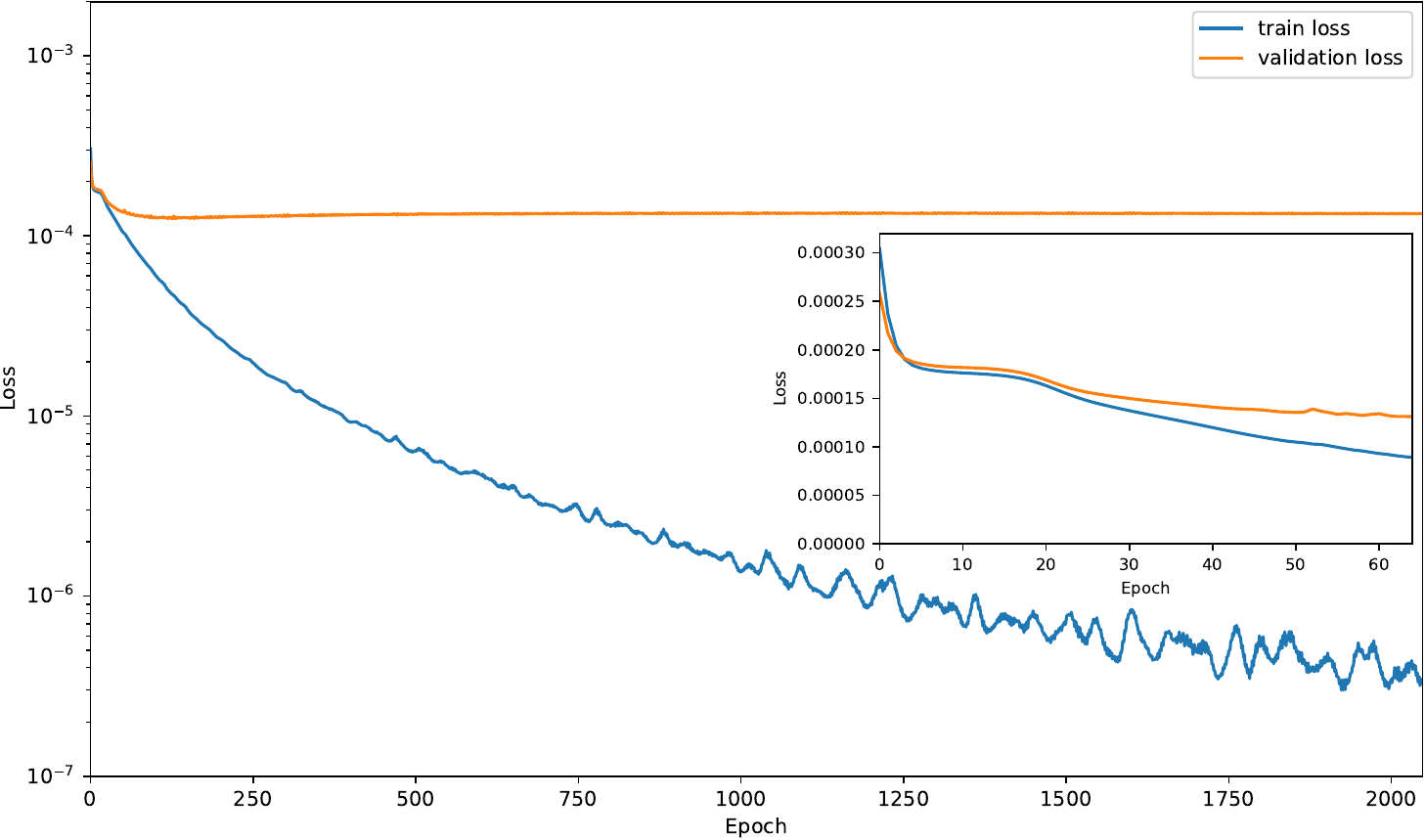}
    \caption{Mean squared error (MSE) plotted for the training and validation sets for the neural network in Figure \ref{openai_embedding_to_embedding_vectorisation_neural_network}.}
    \label{fig:mse_openai_embedding_to_embedding_2048_pdf}
\end{figure}

The main difference to the previous model is that now we use the two embeddings, in two directions. Once we train the neural network to map the input embeddings to output embeddings, we can then predict output embedding(s) for training/validation/test inputs. We then use those output embedding(s) to search for $k$ nearest embeddings neighbours (where $k$ is the number of artworks on the input exhibition, or $k=16$ if not available). Those $k$ nearest embeddings neighbours correspond to $k$ nearest artworks from our large Met Museum database. The training follows the same approach as before, with the training and validation sets split in the same way. The MSE for the model was as follows, as depicted in Figure \ref{fig:mse_openai_embedding_to_embedding_2048_pdf}.

\begin{figure}[htb]
    \centering
    \includegraphics[width=0.75\linewidth]{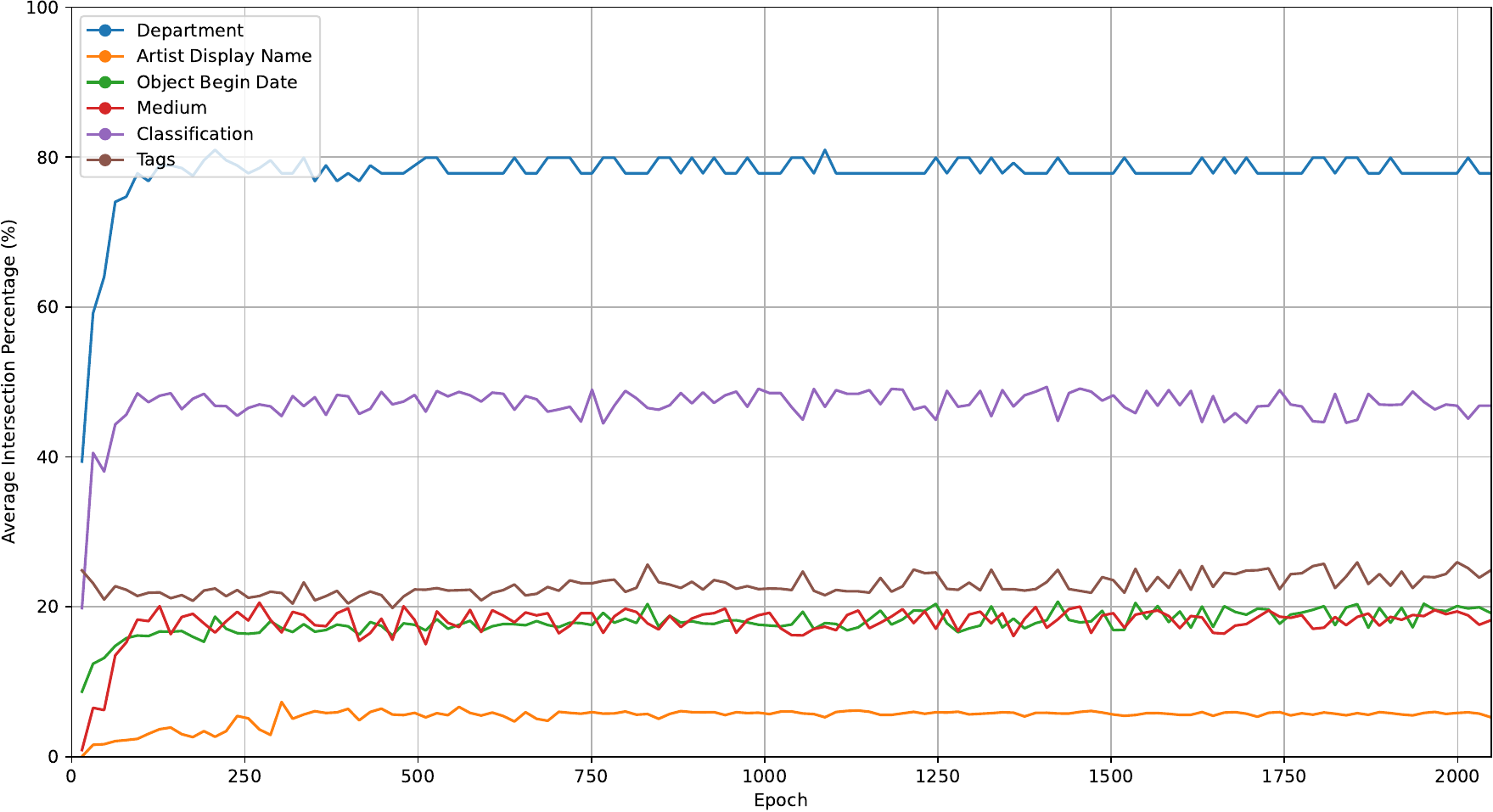}
    \caption{Intersection percentage of the generalised tags per sub-group as a function of epoch within the validation set for the neural network in Figure \ref{openai_embedding_to_embedding_vectorisation_neural_network}.}
    \label{fig:percentage_interception_openai_embedding_to_embedding-crop_2048_pdf}
\end{figure}

The convergence is again quite good, with the saturation for the validation set around 120--150 epochs. This is also reflected in the chart of the intersection percentage of the generalised tags per sub-group as a function of epoch within the validation set in Figure \ref{fig:percentage_interception_openai_embedding_to_embedding-crop_2048_pdf}.

The results for the percentage of intersection predicted versus actual artworks for the validation set is depicted below in Figure \ref{fig:percentage_intersection_artworks_openai_embedding_to_embedding_2048_pdf}. It is higher (at 2048 epochs) than for the models before. We get, as can be seen Figure \ref{fig:percentage_intersection_artworks_openai_embedding_to_embedding_2048_pdf}, of around $4\%$ or below. The value at 2048 epochs is $3.72\%$, so around 400 times better than random choice.

\begin{figure}[htb]
    \centering
    \includegraphics[width=0.75\linewidth]{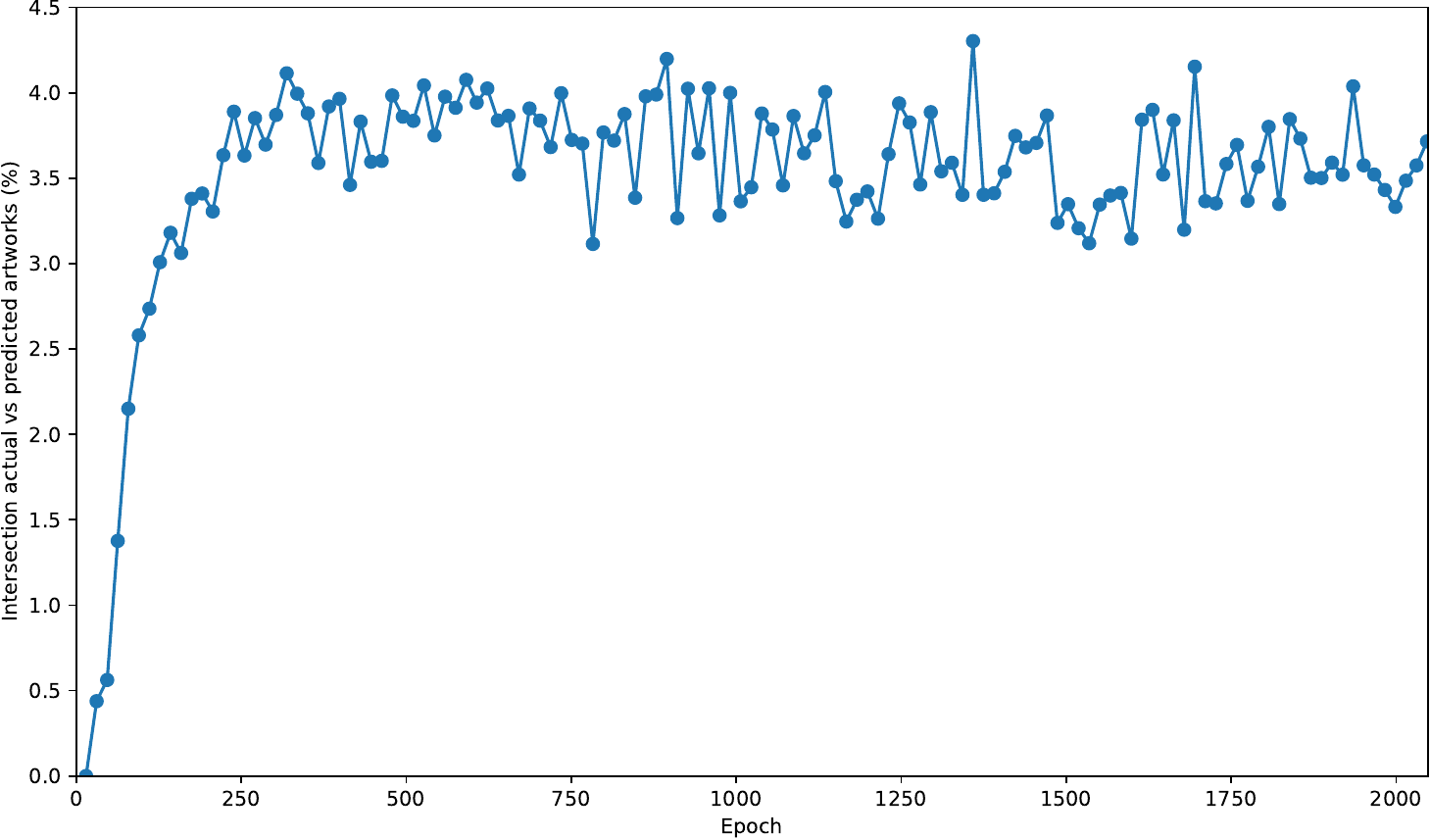}
    \caption{Average percentage of intersection artworks for the validation set for the neural network in Figure \ref{openai_embedding_to_embedding_vectorisation_neural_network}. Again, note that the percentage
overlap based on random choice is around  $0.00907\%$, as demonstrated in the text.}
    \label{fig:percentage_intersection_artworks_openai_embedding_to_embedding_2048_pdf}
\end{figure}

We then test the model out-of-sample, and input the following title and description for a hypothetical exhibition.

\begin{ttquotesmall}
Title of exhibition is: The First Impressionists in Paris and the description is: the paintings of still life from the first Impressionists who created the modern art.
\end{ttquotesmall}

The output was, using 2048 epochs and a number of probes $n_{probe} = 4$ (notice that probes refers to the number of clusters that the  algorithm used by FAISS search use during a query -- the higher the more accuracy we get but less performance/speed):

\begin{table}[htb]
\centering
\tiny
\begin{tabular}{l|l|l|c|c|c}
\hline
\makecell{\textbf{Object}\\ \textbf{ID}}  & \textbf{Department} & \makecell{\textbf{Artist}\\ \textbf{Display Name}} & \makecell{\textbf{Object}\\ \textbf{Begin}\\ \textbf{Date}} & \textbf{Medium} & \textbf{Classification} \\
\hline
693862 & Drawings and Prints & [Goupil et Cie, Braun \& Cie, \ldots] & 1914 & Photogravure & [Prints] \\
356848 & Drawings and Prints & [Edward Ancourt, Gustave Courbet, \ldots] & 1868 & Lithograph with hand coloring & [Prints] \\
489100 & Modern and Contemporary Art & [Jean Metzinger] & 1917 & Oil on canvas & [Paintings] \\
437303 & European Paintings & [Camille Pissarro] & 1880 & Oil on canvas & [Paintings] \\
488853 & Modern and Contemporary Art & [Edouard Vuillard] & 1889 & Oil on canvas & [Paintings] \\
679686 & European Paintings & [Fernand Khnopff] & 1884 & Oil on canvas & [Paintings] \\
483504 & Modern and Contemporary Art & [Georges Braque] & 1923 & Oil and sand on canvas & [Paintings] \\
436529 & European Paintings & [Vincent van Gogh] & 1888 & Oil on canvas & [Paintings] \\
435997 & European Paintings & [Pierre-Auguste Cot] & 1880 & Oil on canvas & [Paintings] \\
388319 & Drawings and Prints & [Jules-Edmond-Charles Lachaise, \ldots] & 1850 & Oil paint on canvas support & [Drawings] \\
438435 & European Paintings & [Claude Monet] & 1872 & Oil on canvas & [Paintings] \\
437347 & European Paintings & [Pierre Puvis de Chavannes] & 1878 & Oil on canvas & [Paintings] \\
490035 & Modern and Contemporary Art & [Edouard Vuillard] & 1897 & Oil on cardboard & [Paintings] \\
438136 & European Paintings & [Paul Cézanne] & 1870 & Oil on canvas & [Paintings] \\
435876 & European Paintings & [Paul Cézanne] & 1888 & Oil on canvas & [Paintings] \\
482211 & Modern and Contemporary Art & [Jacques Villon] & 1905 & Oil on canvas & [Paintings] \\
\hline
\end{tabular}
\end{table}

The marble image for these works of art is depicted below in Figure \ref{marble_openai_embedding_to_embedding_epochs=2048_filter_w_tags_236_training}.
The results in the above table and displayed in Figure \ref{marble_openai_embedding_to_embedding_epochs=2048_filter_w_tags_236_training} seem to show an improvement. 
In particular, we now find some early impressionists, such as Claude Monet and Camille Pissarro. We also have a high 
percentage of intersection of predicted artworks versus actual artworks on the validation set, as depicted in Figure \ref{fig:percentage_intersection_artworks_openai_embedding_to_embedding_2048_pdf}, with an end value at 2048 epochs of around 400 times larger than the one expected by random choice.
In terms of finding instances of still life themes, we find that we get only two still life instances, and furthermore these are from a cubist, Jean Metzinger, a painting from 1917, and from a symbolist, Fernand Khnopff, with a painting from 1884, as can be seen  in Figure \ref{marble_openai_embedding_to_embedding_epochs=2048_filter_w_tags_236_training}. Finally, the time interval of the predicted artworks, 1850--1923, 73 years, is much larger than the length of the impressionist period.  So, this does not seem like a conclusive improvement over the previous simpler model, the one in Section \ref{openai_embedding_vectorisation}. We now attempt using the full-scale power of OpenAI LLM models in the next section.

\begin{figure}[htb]
    \centering
    \includegraphics[width=0.75\linewidth]{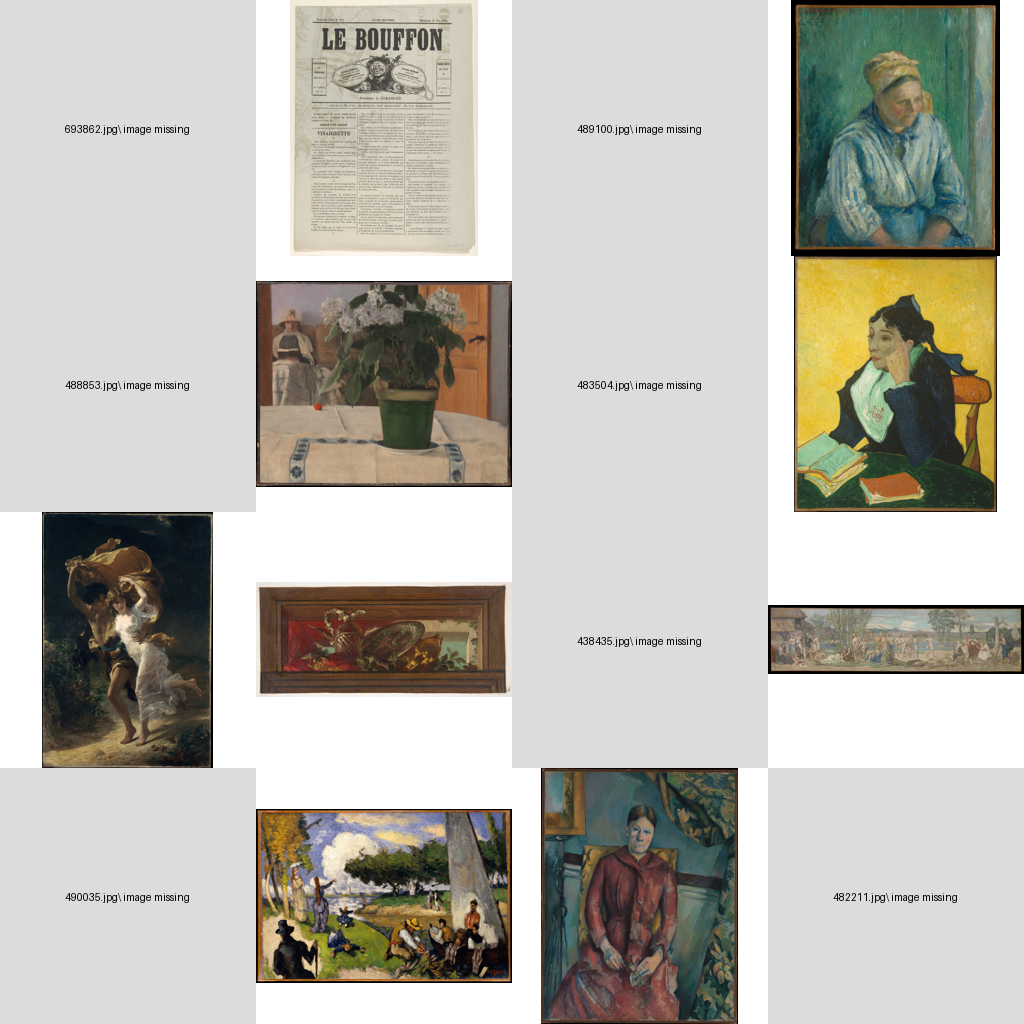}
    \caption{Marble composite of images from the Met Museum corresponding to the run with epochs = 2048 for the model training with the entire dataset, using as input OpenAI embeddings of the title/description of each exhibition and output the OpenAI embeddings of concatenated generalised tags for all artworks in each exhibition. Notice we only display the images of the artworks that are on the public domain, as provided by the Met Museum website and its API \citep{metmuseum_openaccess}. Non-available images are marked by their object id JPEG name.}
	\label{marble_openai_embedding_to_embedding_epochs=2048_filter_w_tags_236_training}
\end{figure}

\subsection{Full OpenAI approach including embedding and mapping}
\label{fine_tuned_openai}

Fourth and finally, we fully use external help, this time by using one of the latest OpenAI GPT (Chat Generative Pre-Trained Transformer models -- \citet{radford2018improving, brown2020language}), in this case 
\texttt{gpt-4o-mini}. These GPT-based models are trained from very large amounts of data scraped from the open Internet, in particular Wikipedia, the arXiv.org e-Print archive, Reddit, among others. These models are a subset of the so-called large language models (LLMs), whereby one maps an input variable $x$, corresponding to the question, to an output variable $y$, corresponding to the answer, using a complex mathematical function, which is encoded in the values of the weights of the nodes of a (very large\footnote{It is estimated that this particular model may have up to 8 billion parameters. However, this is just an estimate as OpenAI usually does not officially disclose exact number of parameters in their models.}) neural network. The transformation of the question, which is a plain text input, into a numerical vector $x$ of very high dimension is similar to the method described in the Sections \ref{self_contained_text_vectorisation}, \ref{openai_embedding_vectorisation} and \ref{openai_embedding_embedding}, an embedding, but it is assumed (since again it is not disclosed by OpenAI) that \texttt{gpt-4o-mini} has its own embedding model. The same goes for the output variable $y$. The mapping is not an explicit analytical equation, but rather a numerical function discovered by the training process using a technique called back-propagation \citep{rumelhart1986learning}, and encoded in the numerical weights of the nodes of a very large neural network. 
In order to use \texttt{gpt-4o-mini} to create a list of artworks for an exhibition, we use the following steps. First, we fine-tune the model, by showing it examples in the form $\langle x, y\rangle$, where $x$ is the input text as in the listing above \ref{all_flat_exhibitions_json} and $y$ is a table of the metadata items for all the artworks presented within the exhibition (see Listing \ref{all_exhibitions_training_set_fine_tuning} for an example). Fine-tuning is a machine learning 
technique whereby one takes an already pre-trained model (such as one of the GPT models) and adapts it to a new domain and/or makes it more specific to a sub-domain \citep{yosinski2014transferable, howard2018universal, radford2018improving, wei2021finetuned}. The fine-tuning process re-trains the last few layers of the neural network, while maintaining most layers intact. This process then improves the model for the new domain and/or sub-domain we are interested in.\footnote{It is well known that there is a kind of feature hierarchy in neural networks, whereby the first layers encode small features while the last layers encode overall generic task or domain specific features, as demonstrated in \citet{yosinski2014transferable}.} A common analogy used is of a very young child learning a second language, the knowledge gained during the first language acquisition is re-used and fine-tuned for the second language. An example would be to retrain the weights between the second hidden layer and the output layer on our example toy model neural network in Figure \ref{fig:conceptual-nn}.

\begin{listing}[htb]
\begin{lstlisting}

{"messages": 
	[
	{"role": "system", "content": "ArtCurator is a factual chatbot that is an expert in JSON format and in artworks and exhibitions from The Metropolitan Museum of Art."}, 

	{"role": "user", "content": "Title of exhibition is: Sculpture and Decorative Arts of the Spanish Renaissance and the description is: The Metropolitan Museum of Art's small but excellent collection of Spanish polychrome sculpture, including sacred reliefs and freestanding carved figures once housed in the churches of Spain, is displayed in the gallery adjacent to the newly reopened Velez Blanco Patio. The selection, which displays the unique blending of early western European and Islamic stylistic and technical influences, emphasizes the diversity in the material culture of Renaissance Spain after the Catholic reconquest by Ferdinand and Isabella."}, 

	{"role": "assistant", "content": "\"{'Department': ['European Sculpture and Decorative Arts', 'European Sculpture and Decorative Arts', 'European Sculpture and Decorative Arts', 'The American Wing', 'European Sculpture and Decorative Arts', 'European Sculpture and Decorative Arts', 'European Sculpture and Decorative Arts', 'European Sculpture and Decorative Arts', 'European Sculpture and Decorative Arts', 'European Sculpture and Decorative Arts', 'European Sculpture and Decorative Arts'], 'Artist Display Name': ['None', 'None', 'None', 'None', 'Diego de Pesquera', 'None', 'None', 'None', 'Juan Martinez Montanes', 'Juan de Ancheta', 'Diego de Atienza'], 'Object Begin Date': ['1600', '1500', '1585', '1630', '1567', '1585', '1585', '1600', '1615', '1575', '1646'], 'Medium': ['Tin-glazed earthenware', 'Tin-glazed and luster-painted earthenware', 'Tin-glazed and luster-painted earthenware', 'Silver gilt, enamel', 'Wood, painted and gilt', 'Wool, silk, metal thread on canvas', 'Wool, silk, metal thread on canvas', 'Tin-glazed and luster-painted earthenware', 'Polychromed wood with gilding', 'Wood, polychromed and gilded', 'Silver gilt with enamel, cast, chased, and engraved'], 'Classification': ['Ceramics-Faience', 'Ceramics-Pottery', 'Ceramics-Pottery', 'None', 'Sculpture', 'Textiles-Embroidered', 'Textiles-Embroidered', 'Ceramics-Pottery', 'Sculpture', 'Sculpture', 'Metalwork-Silver']}\""}
	],
...
}

\end{lstlisting}
\caption{The excerpt of the dataset transformed for the purpose of fine-tuning the GPT model, showing the same exhibition by the Met Museum, as in the previous listings, displaying the role of the GPT model we want it to adopt, the input or user content and the expected GPT model answer, the assistant content.} 
\label{all_exhibitions_training_set_fine_tuning}
\end{listing}

The model used can be pictured as in Figure \ref{fig:openai_fine_tuning_direct_neural_network} below. We fine-tune the model with the examples from the training set (we split the 236 exhibitions in the same way as in previous sections) in the format as in Listing \ref{all_exhibitions_training_set_fine_tuning}. The fine-tuning includes checking the validation set, which is used only for monitoring and evaluation purposes (no automated hyper-parameter optimisation). The training set is used to create a fine-tune GPT model, which can then be called to do evaluation both in the training and validation sets, or on a test set made of out-of-sample examples.

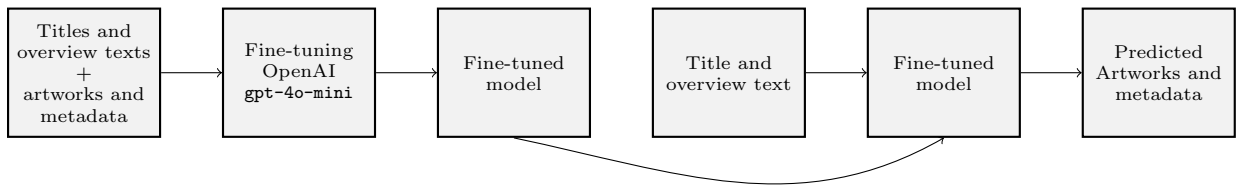
\begin{figure}[htb]
\centering
\scriptsize
\scalebox{0.78}{
\begin{tikzpicture}[
    box/.style={draw, thick, fill=gray!10, minimum width=2.0cm, minimum height=1.7cm, align=center},
    arrow/.style={thick, -{Latex[length=3mm]}},
    node distance=0.8cm and 0.8cm
  ]

  \node[box] (training_set) {Titles and\\overview texts\\+\\artworks and \\metadata};
  \node[box, right=of training_set] (fine_tuning) {Fine-tuning\\OpenAI\\\texttt{gpt-4o-mini}};
  \node[box, right=of fine_tuning] (fine_tuned) {Fine-tuned\\model};
  \node[box, right=of fine_tuned] (input) {Title and\\overview text};
  \node[box, right=of input] (fine_tuned2) {Fine-tuned\\model};
  \node[box, right=of fine_tuned2] (output) {Predicted\\Artworks and\\metadata};

  \draw[->] (training_set) -- (fine_tuning);
  \draw[->] (fine_tuning) -- (fine_tuned);
  \draw[->] (input) -- (fine_tuned2);
  \draw[->] (fine_tuned2) -- (output);
  
  \draw[->] (fine_tuned.south) 
    to[out=-12, in=-150] 
    (fine_tuned2.south);

\end{tikzpicture}
}
\caption{Two-step neural network approach using the fine-tune versions of OpenAI's \texttt{gpt-4o-mini} (version 2024-07-18). The input text is the title/description of each exhibition, and the output is a formatted version of the listing \ref{all_exhibitions_training_set_fine_tuning}.}
\label{fig:openai_fine_tuning_direct_neural_network}
\end{figure}

After we fine-tune the model, in this case \texttt{gpt-4o-mini} (version 2024-07-18), we can use it directly. We can query using inputs within the validation set or out-of-sample test set examples. As before, we split the set of exhibitions we have into a training and a validation set, in the proportion $80\%$ examples for training (a total of 188 examples) and $20\%$ samples for validation or in-sample testing (a total of 48 examples). We notice that the output of the model this time can be of variable length, that is to say, the number of predicted artworks this time is variable (in all the other models before, the selection is done based on the number of actual artworks in each exhibition). This we see as an advantage of this model. We also notice that more  often than not the model outputs a series of artworks with generalised tags in an invalid format (e.g., values missing, rows/columns incomplete, etc.). In these, we simply rerun the model until we get a output that is fully capable of being parsed, having given up trying to fix the invalid formats. We also notice that the output of GPT although it is a list of rows with ``Department'', ``Artist Display Name'', ``Object Begin Date'', ``Medium'', ``Classification'', ``Tags'', these are not necessary either real existing artworks (i.e., inconsistent data can be output, the so-called LLM hallucinations; see e.g., \citet{farquhar2024detecting} and references therein)
 or artworks within the Met Museum. Therefore, we still need to use the Met Museum database to find the nearest artworks to our output, using the same approach with the hit function as defined in Equation \ref{hit_function}. 

\begin{figure}[htb]
    \centering
    \includegraphics[width=0.75\linewidth]{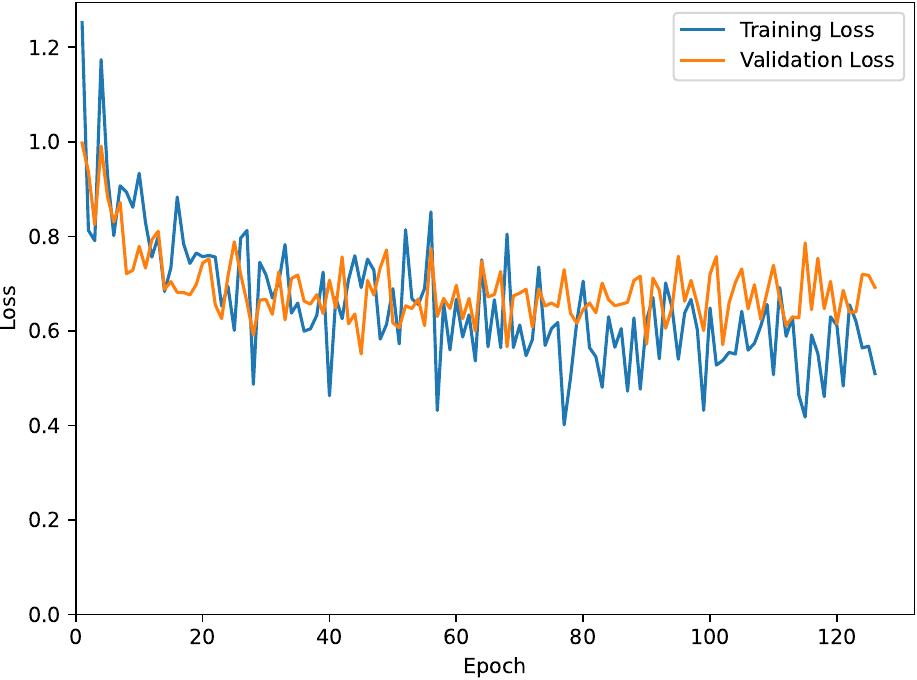}
    \caption{Mean squared error (MSE) plotted for the training and validation sets for the neural network in Figure \ref{fig:openai_fine_tuning_direct_neural_network}. The model was obtained from fine-tuning \texttt{gpt-4o-mini-2024-07-18}, with parameters \texttt{batch\_size=16} and  \texttt{learning\_rate\_multiplier=0.3} and the total number of trained tokens was \num{3036784}. }
    \label{fig:mse_openai_fine_tune}
\end{figure}

We plot the MSE of the training and validation sets for our fine-tuning in Figure \ref{fig:mse_openai_fine_tune}. This time the convergence was not as clear as in the other models, however, this was not a problem as we can see later -- fine-tuning typically changes the model weights only slightly rather than completely refitting the training dataset, so loss (MSE) reduction can be modest, see e.g., \citet[Fig.~4, p.~336]{howard2018universal}.

While before we had the charts of percentage of generalised tags intersection on the validation set, between actual and prediction by the model
and the average of the percentage of predicted artworks that match actual artworks, we notice that OpenAI does not provide 
the intermediate fine-tuned models at each epoch (only at the end), so we could not run those metrics on all intermediate steps. The results at the end, using the validation set are:

\begin{table}[htb]
\centering
\footnotesize
\begin{tabular}{lcc}
\hline
& \begin{tabular}[c]{@{}c@{}}Generalised tag \\ intersection (\%)\end{tabular} 
& \begin{tabular}[c]{@{}c@{}}Predicted artworks \\ matching actual (\%)\end{tabular} \\
\hline
\begin{tabular}[c]{@{}l@{}}Self-contained text \\ vectorisation\end{tabular} & 
{\tiny
\begin{tabular}{lr}
Department & 50.35\% \\
Artist Display Name &3.43\% \\
Object Begin Date & 14.85\% \\
Medium & 11.88\% \\
Classification & 48.21\% \\
Tags & 18.83\%
\end{tabular}
}
 & 0.637\% \\
 \hline
\begin{tabular}[c]{@{}l@{}}OpenAI-based embedding \\ vectorisation\end{tabular} & 
{\tiny
\begin{tabular}{lr}
Department & 80.17\% \\
Artist Display Name &3.28\% \\
Object Begin Date & 20.18\% \\
Medium & 24.68\% \\
Classification & 68.28\% \\
Tags & 26.86\%
\end{tabular}
}
& 1.336\% \\
 \hline
\begin{tabular}[c]{@{}l@{}}Direct NN embedding \\ to artwork metadata\end{tabular} & 
{\tiny
\begin{tabular}{lr}
Department & 77.85\% \\
Artist Display Name & 2.90\% \\
Object Begin Date & 19.92\% \\
Medium & 18.01\% \\
Classification & 42.98\% \\
Tags & 25.65\%
\end{tabular}
}
 & 3.720\% \\
 \hline
\begin{tabular}[c]{@{}l@{}}Full OpenAI approach \\ (embedding + mapping)\end{tabular} & 
{\tiny
\begin{tabular}{lr}
Department & 88.19\% \\
Artist Display Name & 42.88\% \\
Object Begin Date & 31.89\% \\
Medium & 40.47\% \\
Classification & 76.22\% \\
Tags & 22.26\%
\end{tabular}
} & 7.623\% \\
\hline
\end{tabular}
\caption{Comparison of model performance on the validation set based on generalised tag intersection and actual artwork prediction accuracy. For the first, second and third models, we use epochs = 2048, while for the fourth model we use the fine-tuned model as described in the text in Section \ref{fine_tuned_openai}. 
Again, note that the percentage
overlap based on random choice is around  $0.00907\%$, as demonstrated in the text.}
\label{tab:validation_performance}
\end{table}

The results seem to show an improvement in all the generalised tag intersection percentages as well as the percentage of intersection of predicted versus actual artworks. The latter is around $7.623\%$, which is around 800 or so times what one could expect if we did a random choice.

As before, we also test the model out-of-sample, and input the following title and description for a hypothetical exhibition.

\begin{ttquotesmall}
Title of exhibition is: The First Impressionists in Paris and the description is: the paintings of still life from the first Impressionists who created the modern art.
\end{ttquotesmall}

The output was, using the fine-tuned model:

\begin{table}[htb]
\centering
\tiny
\begin{tabular}{l|l|l|c|c|c}
\hline
\makecell{\textbf{Object}\\ \textbf{ID}} & \textbf{Department} & \makecell{\textbf{Artist}\\ \textbf{Display Name}} & \makecell{\textbf{Object}\\ \textbf{Begin Date}} & \textbf{Medium} & \textbf{Classification} \\
\hline
437299 & European Paintings & Camille Pissarro & 1867 & Oil on canvas & Paintings \\
436945 & European Paintings & Edouard Manet & 1862 & Oil on canvas & Paintings \\
437317 & European Paintings & Camille Pissarro & 1872 & Oil on canvas & Paintings \\
436020 & European Paintings & Gustave Courbet & 1873 & Oil on canvas & Paintings \\
436961 & European Paintings & Edouard Manet & 1864 & Oil on canvas & Paintings \\
436952 & European Paintings & Edouard Manet & 1867 & Oil on canvas & Paintings \\
436951 & European Paintings & Edouard Manet & 1862 & Oil on canvas & Paintings \\
436964 & European Paintings & Edouard Manet & 1866 & Oil on canvas & Paintings \\
436013 & European Paintings & Gustave Courbet & 1873 & Oil on canvas & Paintings \\
437303 & European Paintings & Camille Pissarro & 1880 & Oil on canvas & Paintings \\
436960 & European Paintings & Edouard Manet & 1866 & Oil on canvas & Paintings \\
438002 & European Paintings & Edouard Manet & 1880 & Oil on canvas & Paintings \\
438009 & European Paintings & Berthe Morisot & 1865 & Oil on canvas & Paintings \\
436002 & European Paintings & Gustave Courbet & 1866 & Oil on canvas & Paintings \\
436001 & European Paintings & Gustave Courbet & 1865 & Oil on canvas & Paintings \\
435965 & European Paintings & Camille Corot & 1865 & Oil on canvas & Paintings \\
437104 & European Paintings & Claude Monet & 1864 & Oil on canvas & Paintings \\
436150 & European Paintings & Edgar Degas & 1867 & Oil on canvas & Paintings \\
441111 & European Paintings & Michele Gordigiani & 1864 & Oil on canvas & Paintings \\
437436 & European Paintings & Auguste Renoir & 1865 & Oil on canvas & Paintings \\
441104 & European Paintings & Auguste Renoir & 1880 & Oil on canvas & Paintings \\
437097 & European Paintings & Jean-François Millet & 1869 & Oil on canvas & Paintings \\
437094 & European Paintings & Jean-François Millet & 1872 & Oil on canvas & Paintings \\
437107 & European Paintings & Claude Monet & 1876 & Oil on canvas & Paintings \\
441351 & European Paintings & Georges Clairin & 1872 & Oil on canvas & Paintings \\
436519 & European Paintings & Pierre-Paul-Léon Glaize & 1873 & Oil on canvas & Paintings \\
437111 & European Paintings & Claude Monet & 1880 & Oil on canvas & Paintings \\
437109 & European Paintings & Claude Monet & 1880 & Oil on canvas & Paintings \\
437110 & European Paintings & Claude Monet & 1880 & Oil on canvas & Paintings \\
435627 & European Paintings & Charles-Edouard de Beaumont & 1875 & Oil on canvas & Paintings \\
441353 & European Paintings & Benjamin-Constant (Jean-Joseph-Benjamin Constant) & 1876 & Oil on canvas & Paintings \\
435979 & European Paintings & Camille Corot & 1862 & Oil on canvas & Paintings \\
437133 & European Paintings & Claude Monet & 1867 & Oil on canvas & Paintings \\
437135 & European Paintings & Claude Monet & 1869 & Oil on canvas & Paintings \\
437138 & European Paintings & Claude Monet & 1880 & Oil on canvas & Paintings \\
435984 & European Paintings & Camille Corot & 1865 & Oil on canvas & Paintings \\
\hline
\end{tabular}
\end{table}

\begin{figure}[!htb]
    \centering
    \includegraphics[width=0.75\linewidth]{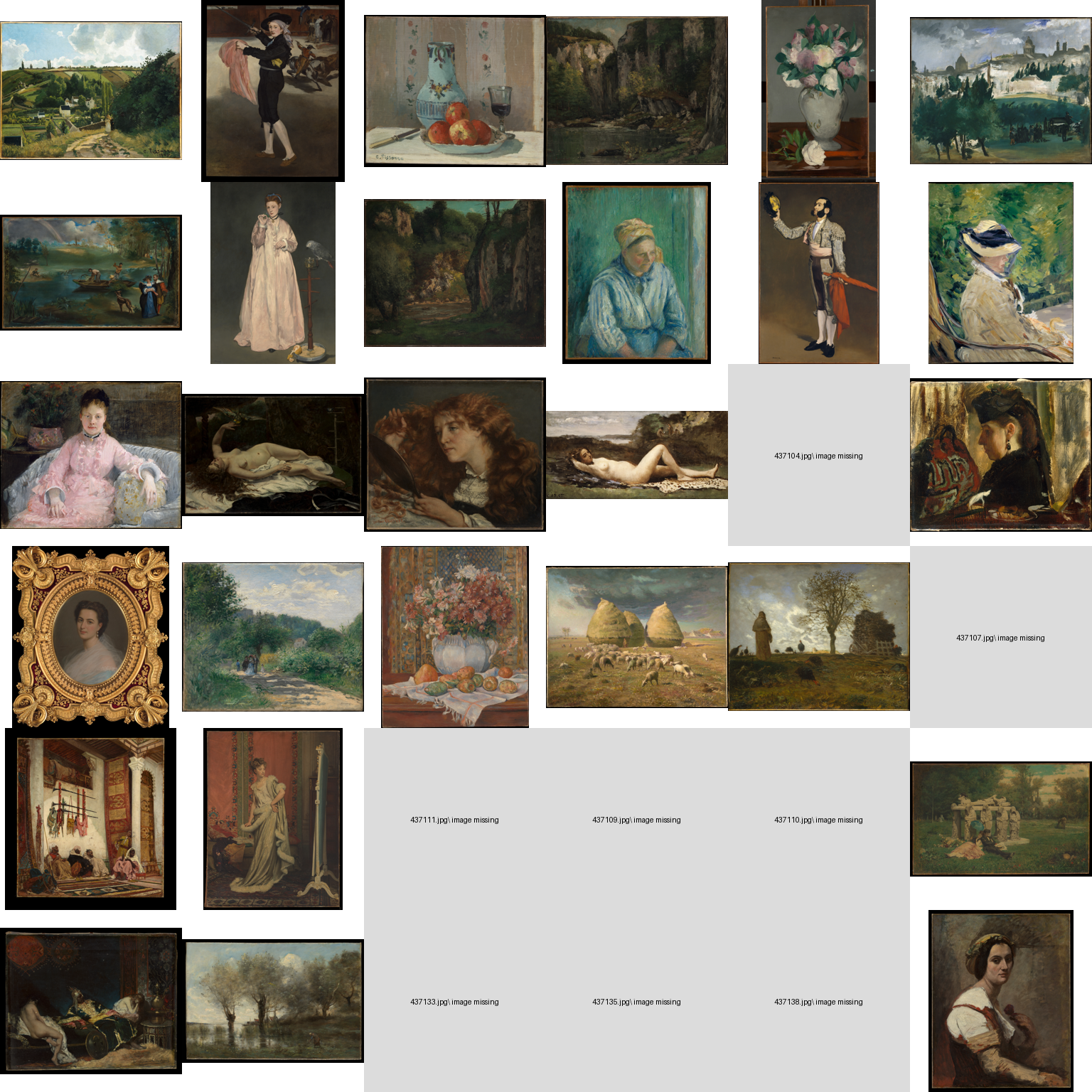}
    \caption{Marble composite of images from the Met Museum corresponding to the run for the fine-tune model trained on the training set only, using as input the text (title/description) of the exhibition and as output the actual list of artworks and their metadata in their original format as in the listing \ref{all_exhibitions_training_set_fine_tuning}. Notice we only display the images of the artworks that are on the public domain, as provided by the Met Museum website and its API \citep{metmuseum_openaccess}. Non-available images are marked by their object id JPEG name.}
	\label{marble_fine_filter_w_tags_188_training}
\end{figure}

The results in the above table and in Figure \ref{marble_fine_filter_w_tags_188_training}, seem to show some, but not all, improvements across all metrics. First, the 
list contains more actual impressionists: Camille Pissarro, Claude Monet, Edgar Degas, Auguste Renoir and Berthe Morisot, so the best result so far. Second, we find 3 instances of still life:``Still Life with Apples and Pitcher'' by Camille Pissarro, a painting from 1872; ``Peonies'' by Edouard Manet from 1864 and ``Still Life with Flowers and Prickly Pears'' by Auguste Renoir from around 1880 or later -- so a result less impressive than the one seen in the model in Section \ref{openai_embedding_vectorisation}, which had 5 instances of still life artworks, and on a smaller predicted set. Third, and finally, the predicted artworks encompass the period span of 1862--1880, 18 years in total, which is a good match to the actual Impressionist period.

\section{Possible further research approaches}

In this article we have focused on four specific approaches, in the above sections, to creating an AI-based curated exhibition. Nonetheless, there are many other ways this research can be expanded and/or enhanced. Here we suggest and review some of those possible avenues of research.

In terms of data, we can think of three main expanded approaches. The first approach one might consider is using more data, from the same data source we used, i.e., the Met Museum. As described in Table \ref{exhibitions_table} we have managed to download a total of 1009 exhibitions from the Met Museum, but we have only used 236 of those. In order to increase this number, we suggest one could do the following:
\begin{enumerate}
\item Collect more input text (the exhibitions' overview descriptions) by hand, as we have automated (via Python) the download of the relevant text. This meant some of them had problems (different formatted text, non-English text, non-ASCII text, image/table interference, etc.) and were removed automatically;
\item Some of the exhibitions were removed because our Python code could not find or could not parse the list of artworks associated with those exhibitions, in particular the artworks object ids which map to the Met Museum's database. In principle one could recover some of those removed exhibitions by doing this work by hand;
\item The Python code was designed so that it could find the tabs/web page fields related to the title, overview text and exhibition objects, but we have noticed the Met Museum have changed several times their web design, not surprisingly, given our dataset spans over a period of 20+ years. Again, one could make an additional effort and parse those rejected or ignored exhibition pages by doing the work by hand.
\end{enumerate}

Second, we could also enhance the number of usable exhibitions by adding those that are composed of artworks that are lent to the Met Museum. However, the reason those were removed is that the Met Museum database does not contain those artworks or theirs associated metadata. To manually add these would be a possible but very difficult and time-consuming task. Third, we could also gather data from other museums, and merge with the dataset we already have from the Met Museum. There are many possibilities, e.g., the Museum of Modern Art \citep{moma_api} provides a collection API; the British Museum has an extensive collection of more than five million objects \citep{britishmuseum2025}; the National Gallery in London also provides an API \citep{ngacuk2025}; and many others. However, this would be a much larger task, and one would have to use as input not just the exhibition test text but also the target museum, as one would need constraints to minimise artworks hosted outside this target museum, to make the exhibition as feasible as possible. Furthermore, each museum uses its own format for the artwork metadata, with different field names/meanings, making the merging of data across multiple sources a very difficult and subjective task.

In terms of modelling itself, we can also think of four different approaches we could use. First, as mentioned in Section \ref{self_contained_text_vectorisation}
we could use a Long Short-Term Memory (LSTM) to turn our task into a variable input/variable output problem. This is similar to what is used in human languages translations, e.g., English text to a French text \citep{sutskever2014sequence}. The advantage of a LSTM neural network is that being a recurrent neural network it would allow for variable input/output sizes, which is part of our problem. While we have gone around this issue in this article by constructing models that transform the problem into a fixed input/output size, there is in principle no actual obstacle in using a LSTM to output a variable list of artworks. However, in practice with such a small amount of training examples (maximum of 236), the data size made this approach unrealistic. Maybe with a much larger dataset, including other museum's data would make it possible. Second, we could construct a slightly different approach as we used in model in Section \ref{openai_embedding_embedding}.
While in that model we created an embedding of the input exhibition title/description and an embedding of the output concatenated list of metadata generalised tags (one string), we could instead unfold our training dataset by mapping each exhibition's embedding to each metadata string embedding, so e.g., if one exhibition had $n$ artworks each with $m$ generalised tags, then we would create in effect $n m$ input/output embeddings. The reason we did not try this is that the resulting graph database and subsequent search would be too large/slow for our purpose, given our computational constrains. Third, 
we could in principle add our own tags by downloading all images of each artwork from the Met Museum database and using a Convolutional Neural Network (CNN) to extract our own extra features/metadata \citep{lecun2002gradient,krizhevsky2012imagenet}. However, this is a massive undertaking, requiring lots of computational power and storage space. Furthermore, this would require inside access as most images are not in the public domain. Of the total \num{484956} works of art, only \num{248472} are in the public domain and available for download. Fourth, and finally, we could also include in our models more of the Met Museum generalised tags. There are 50+ fields that the Met Museum provides in its full artwork metadata set, we just use 6 of those, as we recall these were ``Department'',``Artist Display Name'', ``Object Begin Date'', ``Medium'', ``Classification'', ``Tags''. We did attempt to add other fields, but saw such minor or null improvements, at a computational cost and reduction of simplicity/understanding which we deemed unnecessary.

\section{Conclusion}\label{conclusion}
We have attempted to create four machine learning or artificial intelligence models that learn the behaviour of human art curators. We used data from 25 years of exhibitions at the Metropolitan Museum of Art in New York to train and optimise the models. Three of the four models can be shown to imitate human behaviour to a reasonable accuracy, namely, well above random choice for the set of existing exhibitions and the set of existing artworks in the full collection. Furthermore, as the models get increasingly complex, i.e., as we move from relying simply on the dataset statistical information and start using more and more the knowledge inherent in the so-called large language embedding and query models, we can show we get some improvement in the accuracy, measured by the selection of artworks and its metadata. Nonetheless, the improvement is not very large, and this seems to show that even models with modest size (e.g., our second model with input embeddings and output metadata probabilities) can perform almost as well as very large models using brute force. In particular, this can be seen in our out-of-sample experiment (the one with the impressionists related input text). Finally, we believe that as we gather more data in the future and improve the models, that these kinds of machine learning approaches may be able to get closer and closer to the level of curatorship that humans possess, although it remains a subject of discussion in philosophy of art if there is any artistic value in this. This could be of interest to art galleries who wish to use AI models to autonomously curate their in-house exhibitions and/or to allow the public to access a website or an app that would create virtual exhibitions based on the public's input exhibition title and description. 

\section*{Declarations}

\begin{itemize}
\item Funding: The author and the research presented in this article were fully and wholly self-funded by the author;
\item The author declares no existing conflicts of interest;
\item Availability of the dataset and materials: All datasets used in this article were from publicly available internet sites
 and their sources are referenced and/or are cited in the article.
\end{itemize}


\bibliography{bibliography.bib}

\begin{thebibliography}{79}
\providecommand{\natexlab}[1]{#1}
\providecommand{\url}[1]{\texttt{#1}}
\expandafter\ifx\csname urlstyle\endcsname\relax
  \providecommand{\doi}[1]{doi: #1}\else
  \providecommand{\doi}{doi: \begingroup \urlstyle{rm}\Url}\fi

\bibitem[Abadi et~al.(2016)Abadi, Barham, Chen, Chen, Davis, Dean, Devin,
  Ghemawat, Irving, Isard, et~al.]{abadi2016tensorflow}
Mart{\'\i}n Abadi, Paul Barham, Jianmin Chen, Zhifeng Chen, Andy Davis, Jeffrey
  Dean, Matthieu Devin, Sanjay Ghemawat, Geoffrey Irving, Michael Isard, et~al.
\newblock $\{$TensorFlow$\}$: a system for $\{$Large-Scale$\}$ machine
  learning.
\newblock In \emph{12th USENIX symposium on operating systems design and
  implementation (OSDI 16)}, pages 265--283, 2016.

\bibitem[Andoni and Indyk(2008)]{andoni2008near}
Alexandr Andoni and Piotr Indyk.
\newblock Near-optimal hashing algorithms for approximate nearest neighbor in
  high dimensions.
\newblock \emph{Communications of the ACM}, 51\penalty0 (1):\penalty0 117--122,
  2008.

\bibitem[{ArtLAS Project}(2025)]{artlas2025}
{ArtLAS Project}.
\newblock Basart: An open access global database of exhibition catalogues,
  19th-21th c.
\newblock
  \url{https://artlas.huma-num.fr/en/artlas-bases-de-donnees-en-acces-public/},
  2025.
\newblock Accessed: 2025-01-16.

\bibitem[Bowen et~al.(2020)Bowen, Giannini, Polmeer, Falconer, Miller, and
  Dunn]{bowen2020computational}
Jonathan~P Bowen, Tula Giannini, Gareth Polmeer, Rachel Falconer, Arthur~I
  Miller, and Stuart Dunn.
\newblock Computational culture and {AI}: Challenging human identity and
  curatorial practice.
\newblock \emph{Proceedings of EVA London 2020 (EVA 2020)}, 2020.

\bibitem[{British Museum}(2025)]{britishmuseum2025}
{British Museum}.
\newblock Collection online, 2025.
\newblock URL \url{https://www.britishmuseum.org/collection}.
\newblock Accessed: 2025-06-10.

\bibitem[Brown et~al.(2020)Brown, Mann, Ryder, Subbiah, Kaplan, Dhariwal,
  Neelakantan, Shyam, Sastry, Askell, et~al.]{brown2020language}
Tom Brown, Benjamin Mann, Nick Ryder, Melanie Subbiah, Jared~D Kaplan, Prafulla
  Dhariwal, Arvind Neelakantan, Pranav Shyam, Girish Sastry, Amanda Askell,
  et~al.
\newblock Language models are few-shot learners.
\newblock \emph{Advances in neural information processing systems},
  33:\penalty0 1877--1901, 2020.

\bibitem[Castellano and Vessio(2022)]{castellano2022deep}
Giovanna Castellano and Gennaro Vessio.
\newblock A deep learning approach to clustering visual arts.
\newblock \emph{International Journal of Computer Vision}, 130\penalty0
  (11):\penalty0 2590--2605, 2022.

\bibitem[Cetinic and Grgic(2013)]{cetinic2013automated}
Eva Cetinic and Sonja Grgic.
\newblock Automated painter recognition based on image feature extraction.
\newblock In \emph{Proceedings ELMAR-2013}, pages 19--22. IEEE, 2013.

\bibitem[Cetinic et~al.(2019)Cetinic, Lipic, and Grgic]{cetinic2019deep}
Eva Cetinic, Tomislav Lipic, and Sonja Grgic.
\newblock A deep learning perspective on beauty, sentiment, and remembrance of
  art.
\newblock \emph{IEEE access}, 7:\penalty0 73694--73710, 2019.

\bibitem[Corley and Mihalcea(2005)]{corley2005measuring}
Courtney~D Corley and Rada Mihalcea.
\newblock Measuring the semantic similarity of texts.
\newblock In \emph{Proceedings of the ACL workshop on empirical modeling of
  semantic equivalence and entailment}, pages 13--18, 2005.

\bibitem[Covas(2020)]{covas2020transfer}
Eurico Covas.
\newblock Transfer learning in spatial--temporal forecasting of the solar
  magnetic field.
\newblock \emph{Astronomische Nachrichten}, 341\penalty0 (4):\penalty0
  384--394, 2020.

\bibitem[Covas(2023)]{covas2023named}
Eurico Covas.
\newblock Named entity recognition using {GPT} for identifying comparable
  companies.
\newblock \emph{arXiv preprint arXiv:2307.07420}, 2023.

\bibitem[Devlin et~al.(2019)Devlin, Chang, Lee, and Toutanova]{devlin2019bert}
Jacob Devlin, Ming-Wei Chang, Kenton Lee, and Kristina Toutanova.
\newblock Bert: Pre-training of deep bidirectional transformers for language
  understanding.
\newblock In \emph{Proceedings of the 2019 conference of the North American
  chapter of the association for computational linguistics: human language
  technologies, volume 1 (long and short papers)}, pages 4171--4186, 2019.

\bibitem[Douze et~al.(2024)Douze, Guzhva, Deng, Johnson, Szilvasy, Mazaré,
  Lomeli, Hosseini, and Jégou]{douze2024faiss}
Matthijs Douze, Alexandr Guzhva, Chengqi Deng, Jeff Johnson, Gergely Szilvasy,
  Pierre-Emmanuel Mazaré, Maria Lomeli, Lucas Hosseini, and Hervé Jégou.
\newblock The {Faiss} library.
\newblock \emph{arXiv}, 2024.

\bibitem[Elman(1990)]{elman1990finding}
Jeffrey~L Elman.
\newblock Finding structure in time.
\newblock \emph{Cognitive science}, 14\penalty0 (2):\penalty0 179--211, 1990.

\bibitem[Esser et~al.(2021)Esser, Rombach, and Ommer]{esser2021taming}
Patrick Esser, Robin Rombach, and Bjorn Ommer.
\newblock Taming transformers for high-resolution image synthesis.
\newblock In \emph{Proceedings of the IEEE/CVF conference on computer vision
  and pattern recognition}, pages 12873--12883, 2021.

\bibitem[{Europeana Foundation}(2024)]{europeana2024}
{Europeana Foundation}.
\newblock Europeana: Europe’s digital cultural heritage platform, 2024.
\newblock URL \url{https://www.europeana.eu/en}.
\newblock Accessed: 2025-04-30.

\bibitem[Farquhar et~al.(2024)Farquhar, Kossen, Kuhn, and
  Gal]{farquhar2024detecting}
Sebastian Farquhar, Jannik Kossen, Lorenz Kuhn, and Yarin Gal.
\newblock Detecting hallucinations in large language models using semantic
  entropy.
\newblock \emph{Nature}, 630\penalty0 (8017):\penalty0 625--630, 2024.

\bibitem[Fosset et~al.(2022)Fosset, El-Mennaoui, Rebei, Calligaro, Di~Maria,
  Nguyen-Ban, Rea, Vallade, Vitullo, Zhang, et~al.]{fosset2022docent}
Antoine Fosset, Mohamed El-Mennaoui, Amine Rebei, Paul Calligaro, Elise~Farge
  Di~Maria, H{\'e}l{\`e}ne Nguyen-Ban, Francesca Rea, Marie-Charlotte Vallade,
  Elisabetta Vitullo, Christophe Zhang, et~al.
\newblock Docent: A content-based recommendation system to discover
  contemporary art.
\newblock \emph{arXiv preprint arXiv:2207.05648}, 2022.

\bibitem[Gatys et~al.(2016)Gatys, Ecker, and Bethge]{gatys2016image}
Leon~A Gatys, Alexander~S Ecker, and Matthias Bethge.
\newblock Image style transfer using convolutional neural networks.
\newblock In \emph{Proceedings of the IEEE conference on computer vision and
  pattern recognition}, pages 2414--2423, 2016.

\bibitem[Geman et~al.(1992)Geman, Bienenstock, and Doursat]{geman1992neural}
Stuart Geman, Elie Bienenstock, and Ren{\'e} Doursat.
\newblock Neural networks and the bias/variance dilemma.
\newblock \emph{Neural computation}, 4\penalty0 (1):\penalty0 1--58, 1992.

\bibitem[Giglietto(2024)]{Giglietto_2024}
Fabio Giglietto.
\newblock Evaluating embedding models for clustering italian political news: A
  comparative study of text-embedding-3-large and umberto, August 2024.
\newblock URL \url{http://dx.doi.org/10.31219/osf.io/2j9ed}.

\bibitem[Gombrich(2023)]{Gombrich2023-gh}
E~H Gombrich.
\newblock \emph{The story of art}.
\newblock Phaidon Press, London, England, April 2023.

\bibitem[Gordon(2025)]{newsobserver2025}
Brian Gordon.
\newblock Meet the new curator at the duke art museum. it isn't human.
\newblock
  \url{https://www.newsobserver.com/entertainment/arts-culture/article279310784.html},
  2025.
\newblock Accessed: 2025-01-16.

\bibitem[Hamilton et~al.(2021)Hamilton, Fu, Lu, Bui, Bopp, Chen, Tran, Wang,
  Rogers, Zhang, et~al.]{hamilton2021mosaic}
Mark Hamilton, Stephanie Fu, Mindren Lu, Johnny Bui, Darius Bopp, Zhenbang
  Chen, Felix Tran, Margaret Wang, Marina Rogers, Lei Zhang, et~al.
\newblock Mosaic: Finding artistic connections across culture with conditional
  image retrieval.
\newblock In \emph{NeurIPS 2020 Competition and Demonstration Track}, pages
  133--155. PMLR, 2021.

\bibitem[He et~al.(2016)He, Fang, Wang, and McAuley]{he2016vista}
Ruining He, Chen Fang, Zhaowen Wang, and Julian McAuley.
\newblock Vista: A visually, socially, and temporally-aware model for artistic
  recommendation.
\newblock In \emph{Proceedings of the 10th ACM conference on recommender
  systems}, pages 309--316, 2016.

\bibitem[Herbert(1988)]{herbert1988impressionism}
Robert~L Herbert.
\newblock \emph{Impressionism: art, leisure, and Parisian society}.
\newblock Yale University Press, 1988.

\bibitem[Herman and Moruzzi(2024)]{10.1162/leon_a_02561}
Laura~M. Herman and Caterina Moruzzi.
\newblock The algorithmic pedestal: A practice-based study of algorithmic and
  artistic curation.
\newblock \emph{Leonardo}, 57\penalty0 (5):\penalty0 485--492, 10 2024.
\newblock ISSN 0024-094X.
\newblock \doi{10.1162/leon_a_02561}.
\newblock URL \url{https://doi.org/10.1162/leon\_a\_02561}.

\bibitem[Hochreiter and Schmidhuber(1997)]{hochreiter1997long}
Sepp Hochreiter and J{\"u}rgen Schmidhuber.
\newblock Long short-term memory.
\newblock \emph{Neural computation}, 9\penalty0 (8):\penalty0 1735--1780, 1997.

\bibitem[Howard and Ruder(2018)]{howard2018universal}
Jeremy Howard and Sebastian Ruder.
\newblock Universal language model fine-tuning for text classification.
\newblock \emph{arXiv preprint arXiv:1801.06146}, 2018.

\bibitem[Hu et~al.(2008)Hu, Koren, and Volinsky]{hu2008collaborative}
Yifan Hu, Yehuda Koren, and Chris Volinsky.
\newblock Collaborative filtering for implicit feedback datasets.
\newblock In \emph{2008 Eighth IEEE international conference on data mining},
  pages 263--272. Ieee, 2008.

\bibitem[{IIIF Consortium}(2024)]{iiif2024}
{IIIF Consortium}.
\newblock International image interoperability framework (iiif), 2024.
\newblock URL \url{https://iiif.io/}.
\newblock Accessed: 2025-04-30.

\bibitem[Iscen et~al.(2017)Iscen, Furon, Gripon, Rabbat, and
  J{\'e}gou]{iscen2017memory}
Ahmet Iscen, Teddy Furon, Vincent Gripon, Michael Rabbat, and Herv{\'e}
  J{\'e}gou.
\newblock Memory vectors for similarity search in high-dimensional spaces.
\newblock \emph{IEEE transactions on big data}, 4\penalty0 (1):\penalty0
  65--77, 2017.

\bibitem[Kenton and Toutanova(2019)]{kenton2019bert}
Jacob Devlin Ming-Wei~Chang Kenton and Lee~Kristina Toutanova.
\newblock Bert: Pre-training of deep bidirectional transformers for language
  understanding.
\newblock In \emph{Proceedings of naacL-HLT}, volume~1, page~2. Minneapolis,
  Minnesota, 2019.

\bibitem[Keraghel et~al.(2024)Keraghel, Morbieu, and Nadif]{keraghel2024beyond}
Imed Keraghel, Stanislas Morbieu, and Mohamed Nadif.
\newblock Beyond words: a comparative analysis of llm embeddings for effective
  clustering.
\newblock In \emph{International Symposium on Intelligent Data Analysis}, pages
  205--216. Springer, 2024.

\bibitem[Keren(2002)]{keren2002painter}
Daniel Keren.
\newblock Painter identification using local features and naive bayes.
\newblock In \emph{2002 International Conference on Pattern Recognition},
  volume~2, pages 474--477. IEEE, 2002.

\bibitem[Kingma(2014)]{kingma2014adam}
Diederik~P Kingma.
\newblock Adam: A method for stochastic optimization.
\newblock \emph{arXiv preprint arXiv:1412.6980}, 2014.

\bibitem[Krizhevsky et~al.(2012)Krizhevsky, Sutskever, and
  Hinton]{krizhevsky2012imagenet}
Alex Krizhevsky, Ilya Sutskever, and Geoffrey~E Hinton.
\newblock Imagenet classification with deep convolutional neural networks.
\newblock \emph{Advances in neural information processing systems}, 25, 2012.

\bibitem[Krysa and Moscoso(2019)]{krysa2019next}
Joasia Krysa and Manuela Moscoso.
\newblock The next biennial should be curated by a machine: A research
  proposition.
\newblock \emph{e-flux}, 2019.

\bibitem[LeCun et~al.(2002)LeCun, Bottou, Bengio, and
  Haffner]{lecun2002gradient}
Yann LeCun, L{\'e}on Bottou, Yoshua Bengio, and Patrick Haffner.
\newblock Gradient-based learning applied to document recognition.
\newblock \emph{Proceedings of the IEEE}, 86\penalty0 (11):\penalty0
  2278--2324, 2002.

\bibitem[Li(2025)]{li2025enhanced}
Weiwei Li.
\newblock Enhanced automated art curation using supervised modified cnn for art
  style classification.
\newblock \emph{Scientific Reports}, 15\penalty0 (1):\penalty0 7319, 2025.

\bibitem[Messina et~al.(2019)Messina, Dominguez, Parra, Trattner, and
  Soto]{messina2019content}
Pablo Messina, Vicente Dominguez, Denis Parra, Christoph Trattner, and Alvaro
  Soto.
\newblock Content-based artwork recommendation: integrating painting metadata
  with neural and manually-engineered visual features.
\newblock \emph{User Modeling and User-Adapted Interaction}, 29\penalty0
  (2):\penalty0 251--290, 2019.

\bibitem[Messina et~al.(2020)Messina, Cartagena, Cerda-Mardini, del Rio, and
  Parra]{messina2020curatornet}
Pablo Messina, Manuel Cartagena, Patricio Cerda-Mardini, Felipe del Rio, and
  Denis Parra.
\newblock Curatornet: Visually-aware recommendation of art images.
\newblock \emph{arXiv preprint arXiv:2009.04426}, 2020.

\bibitem[Mikolov et~al.(2013)Mikolov, Chen, Corrado, and
  Dean]{DBLP:journals/corr/abs-1301-3781}
Tom{\'{a}}s Mikolov, Kai Chen, Greg Corrado, and Jeffrey Dean.
\newblock Efficient estimation of word representations in vector space.
\newblock In Yoshua Bengio and Yann LeCun, editors, \emph{1st International
  Conference on Learning Representations, {ICLR} 2013, Scottsdale, Arizona,
  USA, May 2-4, 2013, Workshop Track Proceedings}, 2013.
\newblock URL \url{http://arxiv.org/abs/1301.3781}.

\bibitem[{Museum of Modern Art (MoMA)}(2025)]{moma2025}
{Museum of Modern Art (MoMA)}.
\newblock Exhibition history.
\newblock \url{https://www.moma.org/calendar/exhibitions/history/?locale=en},
  2025.
\newblock Accessed: 2025-01-16.

\bibitem[{Nasher Museum of Art at Duke University}(2023)]{nasher2023act}
{Nasher Museum of Art at Duke University}.
\newblock Act as if you are a curator: An {AI}-generated exhibition.
\newblock
  \url{https://nasher.duke.edu/exhibitions/act-as-if-you-are-a-curator-an-ai-generated-exhibition/},
  2023.
\newblock Accessed: 2025-01-16.

\bibitem[{National Gallery of Art}(2025)]{nga2025}
{National Gallery of Art}.
\newblock Collection search.
\newblock \url{https://www.nga.gov/collection/collection-search.html}, 2025.
\newblock Accessed: 2025-01-16.

\bibitem[Ohm(2023)]{ohm2023algorithmic}
Tillmann Ohm.
\newblock Algorithmic exhibition-making.
\newblock \emph{AI in Museums}, pages 209--16, 2023.

\bibitem[OpenAI(2023)]{DBLP:journals/corr/abs-2303-08774}
OpenAI.
\newblock {GPT-4} technical report.
\newblock \emph{CoRR}, abs/2303.08774, 2023.
\newblock \doi{10.48550/ARXIV.2303.08774}.
\newblock URL \url{https://doi.org/10.48550/arXiv.2303.08774}.

\bibitem[OpenAI(2024)]{openai2024embedding}
OpenAI.
\newblock New embedding models and {API} updates, 2024.
\newblock URL
  \url{https://openai.com/index/new-embedding-models-and-api-updates/}.
\newblock Accessed: 2025-04-24.

\bibitem[Petukhova et~al.(2025)Petukhova, Matos-Carvalho, and
  Fachada]{petukhova2025text}
Alina Petukhova, Jo{\~a}o~P Matos-Carvalho, and Nuno Fachada.
\newblock Text clustering with large language model embeddings.
\newblock \emph{International Journal of Cognitive Computing in Engineering},
  6:\penalty0 100--108, 2025.

\bibitem[Radford et~al.(2018)Radford, Narasimhan, Salimans, Sutskever,
  et~al.]{radford2018improving}
Alec Radford, Karthik Narasimhan, Tim Salimans, Ilya Sutskever, et~al.
\newblock Improving language understanding by generative pre-training.
\newblock \emph{Google preprint}, 2018.

\bibitem[Radford et~al.(2021)Radford, Kim, Hallacy, Ramesh, Goh, Agarwal,
  Sastry, Askell, Mishkin, Clark, et~al.]{radford2021learning}
Alec Radford, Jong~Wook Kim, Chris Hallacy, Aditya Ramesh, Gabriel Goh,
  Sandhini Agarwal, Girish Sastry, Amanda Askell, Pamela Mishkin, Jack Clark,
  et~al.
\newblock Learning transferable visual models from natural language
  supervision.
\newblock In \emph{International conference on machine learning}, pages
  8748--8763. PmLR, 2021.

\bibitem[Ramesh et~al.(2021)Ramesh, Pavlov, Goh, Gray, Voss, Radford, Chen, and
  Sutskever]{ramesh2021zero}
Aditya Ramesh, Mikhail Pavlov, Gabriel Goh, Scott Gray, Chelsea Voss, Alec
  Radford, Mark Chen, and Ilya Sutskever.
\newblock Zero-shot text-to-image generation.
\newblock In \emph{International conference on machine learning}, pages
  8821--8831. Pmlr, 2021.

\bibitem[Rumelhart et~al.(1986)Rumelhart, Hinton, and
  Williams]{rumelhart1986learning}
David~E Rumelhart, Geoffrey~E Hinton, and Ronald~J Williams.
\newblock Learning representations by back-propagating errors.
\newblock \emph{nature}, 323\penalty0 (6088):\penalty0 533--536, 1986.

\bibitem[Saleh and Elgammal(2015)]{saleh2015large}
Babak Saleh and Ahmed Elgammal.
\newblock Large-scale classification of fine-art paintings: Learning the right
  metric on the right feature.
\newblock \emph{arXiv preprint arXiv:1505.00855}, 2015.

\bibitem[Saleh et~al.(2016)Saleh, Abe, Arora, and Elgammal]{saleh2016toward}
Babak Saleh, Kanako Abe, Ravneet~Singh Arora, and Ahmed Elgammal.
\newblock Toward automated discovery of artistic influence.
\newblock \emph{Multimedia Tools and Applications}, 75:\penalty0 3565--3591,
  2016.

\bibitem[Salton et~al.(1975)Salton, Wong, and Yang]{salton1975vector}
Gerard Salton, A.~Wong, and C.~S. Yang.
\newblock A vector space model for automatic indexing.
\newblock \emph{Communications of the ACM}, 18\penalty0 (11):\penalty0
  613--620, 1975.

\bibitem[Schaerf et~al.(2023)Schaerf, Ballesteros, Bernasconi, Neri, and del
  Castillo]{schaerf2023ai}
Ludovica Schaerf, Pepe Ballesteros, Valentine Bernasconi, Iacopo Neri, and
  Dar{\'\i}o~Negueruela del Castillo.
\newblock {AI} art curation: Re-imagining the city of helsinki in occasion of
  its biennial.
\newblock \emph{arXiv preprint arXiv:2306.03753}, 2023.

\bibitem[Shamir and Tarakhovsky(2012)]{shamir2012computer}
Lior Shamir and Jane~A Tarakhovsky.
\newblock Computer analysis of art.
\newblock \emph{Journal on Computing and Cultural Heritage (JOCCH)}, 5\penalty0
  (2):\penalty0 1--11, 2012.

\bibitem[Shamir et~al.(2010)Shamir, Macura, Orlov, Eckley, and
  Goldberg]{shamir2010impressionism}
Lior Shamir, Tomasz Macura, Nikita Orlov, D~Mark Eckley, and Ilya~G Goldberg.
\newblock Impressionism, expressionism, surrealism: Automated recognition of
  painters and schools of art.
\newblock \emph{ACM Transactions on Applied Perception (TAP)}, 7\penalty0
  (2):\penalty0 1--17, 2010.

\bibitem[Shen et~al.(2019)Shen, Efros, and Aubry]{shen2019discovering}
Xi~Shen, Alexei~A Efros, and Mathieu Aubry.
\newblock Discovering visual patterns in art collections with
  spatially-consistent feature learning.
\newblock In \emph{Proceedings of the IEEE/CVF conference on computer vision
  and pattern recognition}, pages 9278--9287, 2019.

\bibitem[Srinivasan(2024)]{srinivasan2024see}
Ramya Srinivasan.
\newblock To see or not to see: Understanding the tensions of algorithmic
  curation for visual arts.
\newblock In \emph{The 2024 ACM Conference on Fairness, Accountability, and
  Transparency}, pages 444--455, 2024.

\bibitem[Sutskever et~al.(2014)Sutskever, Vinyals, and
  Le]{sutskever2014sequence}
Ilya Sutskever, Oriol Vinyals, and Quoc~V Le.
\newblock Sequence to sequence learning with neural networks.
\newblock \emph{Advances in neural information processing systems}, 27, 2014.

\bibitem[{The Metropolitan Museum of
  Art}(2025{\natexlab{a}})]{metmuseum_openaccess}
{The Metropolitan Museum of Art}.
\newblock The {M}etropolitan {M}useum of {A}rt {C}ollection {API}.
\newblock \url{https://github.com/metmuseum/openaccess}, 2025{\natexlab{a}}.
\newblock Accessed: 2025-01-27.

\bibitem[{The Metropolitan Museum of
  Art}(2025{\natexlab{b}})]{metmuseum_past_exhibitions}
{The Metropolitan Museum of Art}.
\newblock Past exhibitions, 2025{\natexlab{b}}.
\newblock URL \url{https://www.metmuseum.org/exhibitions/past}.
\newblock Accessed: 2025-02-24.

\bibitem[{The Museum of Modern Art}(2025{\natexlab{a}})]{moma_api}
{The Museum of Modern Art}.
\newblock {MoMA} collection {API}.
\newblock \url{https://api.moma.org/}, 2025{\natexlab{a}}.
\newblock URL \url{https://api.moma.org/}.
\newblock Accessed: 2025-06-10.

\bibitem[{The Museum of Modern
  Art}(2025{\natexlab{b}})]{moma_identifying_art_2025}
{The Museum of Modern Art}.
\newblock Identifying art through machine learning.
\newblock
  \url{https://www.moma.org/calendar/exhibitions/history/identifying-art},
  2025{\natexlab{b}}.
\newblock Accessed: 23 June 2025.

\bibitem[{The National Gallery}(2025)]{ngacuk2025}
{The National Gallery}.
\newblock Collection data, 2025.
\newblock URL
  \url{https://www.nationalgallery.org.uk/documentation/ngacuk/collection-data}.
\newblock Accessed: 2025-06-10.

\bibitem[{The Royal Swedish Academy of Sciences}(2024)]{nobel2024press}
{The Royal Swedish Academy of Sciences}.
\newblock The {N}obel {P}rize in {P}hysics 2024.
\newblock \url{https://www.nobelprize.org/prizes/physics/2024/press-release/},
  2024.
\newblock Accessed: 2025-01-16.

\bibitem[{University of Vienna}(2025)]{univie2025}
{University of Vienna}.
\newblock Database of modern exhibitions ({DoME}).
\newblock \url{https://exhibitions.univie.ac.at}, 2025.
\newblock Accessed: 2025-01-16.

\bibitem[Vaswani et~al.(2017)Vaswani, Shazeer, Parmar, Uszkoreit, Jones, Gomez,
  Kaiser, and Polosukhin]{vaswani2017attention}
Ashish Vaswani, Noam Shazeer, Niki Parmar, Jakob Uszkoreit, Llion Jones,
  Aidan~N Gomez, {\L}ukasz Kaiser, and Illia Polosukhin.
\newblock Attention is all you need.
\newblock \emph{Advances in neural information processing systems}, 30, 2017.

\bibitem[Vinyals et~al.(2015)Vinyals, Toshev, Bengio, and
  Erhan]{vinyals2015show}
Oriol Vinyals, Alexander Toshev, Samy Bengio, and Dumitru Erhan.
\newblock Show and tell: A neural image caption generator.
\newblock In \emph{Proceedings of the IEEE conference on computer vision and
  pattern recognition}, pages 3156--3164, 2015.

\bibitem[von Davier et~al.(2024)von Davier, Herman, and
  Moruzzi]{von2024machine}
Thomas~{\c{S}}erban von Davier, Laura~M Herman, and Caterina Moruzzi.
\newblock A machine walks into an exhibit: a technical analysis of art
  curation.
\newblock \emph{Arts}, 13\penalty0 (5):\penalty0 138, 2024.

\bibitem[Wei et~al.(2021)Wei, Bosma, Zhao, Guu, Yu, Lester, Du, Dai, and
  Le]{wei2021finetuned}
Jason Wei, Maarten Bosma, Vincent~Y Zhao, Kelvin Guu, Adams~Wei Yu, Brian
  Lester, Nan Du, Andrew~M Dai, and Quoc~V Le.
\newblock Finetuned language models are zero-shot learners.
\newblock \emph{arXiv preprint arXiv:2109.01652}, 2021.

\bibitem[Yilma and Leiva(2023)]{yilma2023elements}
Bereket~A Yilma and Luis~A Leiva.
\newblock The elements of visual art recommendation: Learning latent semantic
  representations of paintings.
\newblock In \emph{Proceedings of the 2023 CHI Conference on Human Factors in
  Computing Systems}, pages 1--17, 2023.

\bibitem[Yosinski et~al.(2014)Yosinski, Clune, Bengio, and
  Lipson]{yosinski2014transferable}
Jason Yosinski, Jeff Clune, Yoshua Bengio, and Hod Lipson.
\newblock How transferable are features in deep neural networks?
\newblock \emph{Advances in neural information processing systems}, 27, 2014.

\bibitem[Young et~al.(2018)Young, Hazarika, Poria, and
  Cambria]{young2018recent}
Tom Young, Devamanyu Hazarika, Soujanya Poria, and Erik Cambria.
\newblock Recent trends in deep learning based natural language processing.
\newblock \emph{IEEE Computational intelligenCe magazine}, 13\penalty0
  (3):\penalty0 55--75, 2018.

\bibitem[Zhu et~al.(2017)Zhu, Park, Isola, and Efros]{zhu2017unpaired}
Jun-Yan Zhu, Taesung Park, Phillip Isola, and Alexei~A Efros.
\newblock Unpaired image-to-image translation using cycle-consistent
  adversarial networks.
\newblock In \emph{Proceedings of the IEEE international conference on computer
  vision}, pages 2223--2232, 2017.

\end{thebibliography}

\end{document}